\newif\ifARXIV

\newif\ifHAL
\HALtrue

\ifHAL
\documentclass[11pt,a4paper]{article}
\usepackage{natbib}
\usepackage{a4wide}
\else
\documentclass[preprint]{elsarticle}
\fi

\usepackage{moreverb,url}

\usepackage[colorlinks,bookmarksopen,bookmarksnumbered,allcolors=blue]{hyperref}
\ifHAL
\usepackage{doi} 
\fi
\usepackage[ruled,vlined]{algorithm2e}
\SetAlFnt{\scriptsize} 

\usepackage{amsmath,amssymb,amsfonts}
\usepackage[utf8]{inputenc}
\usepackage{graphicx}
\usepackage[scriptsize]{subfigure}
\usepackage{xspace}
\usepackage{genom}
\usepackage{listings}
\usepackage{lstlang3}
\usepackage{xcolor}
\usepackage{makecell}
\usepackage{placeins}
\usepackage{siunitx}

\newcommand{\fiacre}{{\sc Fiacre}}
\newcommand{\hfiacre}{{\sc H-Fiacre}}
\newcommand{\tina}{{\sc Tina}}
\newcommand{\hippo}{{\sc Hippo}}
\newcommand{\ps}{{\sc ProSkill}}
\newcommand{\proskill}{{\sc ProSkill}}
\newcommand{\ProSkill}{{\sc ProSkill}}

\newcommand{\task}[1]{\textsl{#1}}
\newcommand{\event}[1]{\textsl{#1}}
\newcommand{\port}[1]{\textsf{#1}}
\newcommand{\modu}[1]{\textsc{#1}}
\newcommand{\skill}[1]{\textbf{#1}}
\newcommand{\process}[1]{\textbf{#1}}
\newcommand{\sv}[1]{\textit{#1}}
\newcommand{\svv}[1]{\textsf{#1}}
\newcommand{\state}[1]{\textit{#1}}

\newcommand{\code}[1]{{\small\ttfamily{#1}}}
\newcommand{\tool}[1]{\textsf{#1}}
\newcommand{\field}[1]{\code{:#1}}

\usepackage{inconsolata}
\definecolor{eclipseBlue}{RGB}{42,0.0,255}
\definecolor{eclipseGreen}{RGB}{63,127,95}
\definecolor{eclipsePurple}{RGB}{127,0,85}

\definecolor{colKeys}{rgb}{1,0,0.75}
\definecolor{colKeys2}{rgb}{0,0,1}
\definecolor{colIdentifier}{rgb}{0,0,0}
\definecolor{colComments}{rgb}{0,0.5,0}
\definecolor{colString}{rgb}{0.6,0.1,0.1}

\definecolor{bluekeywords}{rgb}{0.13,0.13,1}
\definecolor{greencomments}{rgb}{0,0.5,0}
\definecolor{redstrings}{rgb}{0.9,0,0}

\begin{document}

\renewcommand{\cite}[1]{\citep{#1}}

\ifHAL
\title{\ProSkill{}: A formal skill language for\\acting in robotics\footnote{Part of this work has been supported by the HumFleet project
    ANR-23-CE33-0003 and the EU-funded project euROBIN under grant agreement
no. 101070596 (https://www.eurobin-project.eu/).\\ 
}}
\fi

\ifHAL
\author{Félix Ingrand\\
  felix@laas.fr\\
LAAS-CNRS, Universit\'e de Toulouse, Toulouse, France}
\date{}

\maketitle

\fi

\begin{abstract}
  Acting is an important decisional function for autonomous robots. Acting relies on skills to implement and to model the activities it
  oversees: refinement, local recovery, temporal dispatching, external asynchronous events, and commands execution, all done online. While
  sitting between planning and the robotic platform, acting often relies on programming primitives and an interpreter which executes these
  skills. Following our experience in providing a formal framework to program the functional components of our robots, we propose a new
  language, to program the acting skills. This language maps unequivocally into a formal model which can then be used to check properties
  offline or execute the skills, or more precisely their formal equivalent, and perform runtime verification. We illustrate with a real
  example how we can program a survey mission for a drone in this new language, prove some formal properties on the program and directly
  execute the formal model on the drone to perform the mission.
\end{abstract}

\ifHAL
\fi
\section{Introduction and motivation}
\label{sec:intro}

\label{sec:actingproskill}

\emph{Acting} is an important decisional function to ensure proper deliberation on an autonomous system ~\citep{Ingrand:2015ue}. It often
sits between \emph{planning} and the platform, but unlike \emph{planning} it is an online process, which must stay reactive to the dynamic
of the environment and the platform and cannot devote resources to long computations and complex searches. \emph{Acting} often relies on
models, called \emph{skills}, which specify how to perform actions (as an operational model), while the \emph{action} models used for
\emph{planning} are more what is abstractly needed to perform the action (as a descriptive model)~\cite{Ghallab:2016uo}.

The most basic skills need to connect to the commands made available by the functional level to the \emph{acting} component, call them
asynchronously, get execution status and result, but it also needs means to receive exogenous events as they occur in the environment. This
action/command dispatching may also rely on preconditions and invariants checking, interruptions, temporal constraints, etc.  Above the
basic skills one often finds more complex skills, similar to programs with control structures to allow for local choices and local recoveries with
test, branching, looping, parallel and asynchronous execution.

Considering the expected functionalities of an acting component, its skill language/framework should provide the following features:
\begin{itemize}
\item Support for Validation and Verification (V\&V). Notwithstanding the other functionalities,
  this is the feature the work presented in this paper focuses on.
\item One cannot only rely on basic skills connecting to the robot commands, one also needs some programming primitives (e.g., test,
  branching, loop, parallelism).
\item Controlling and managing multiple commands sent to functional components require some primitives to handle parallelism, threads,
  resources.  
\item In embedded real-time systems, one needs an explicit time representation to wait, synchronize, check temporal
  constraints, etc.
\item  The language should reflect the tight coupling and necessary consistency between \emph{planning}  and \emph{acting}.
\end{itemize}

There are other less critical features which are now expected from an acting component: human awareness (how does the acting component
manage the human presence and activities),  link with motion/manipulation (most robots move and solving their motion planning
problems often require an elaborate coupling with acting).  Acting skills can be programmed by hand, but learning them or their parameters
must also be investigated.

Nevertheless, all these considered, the work presented here focuses on the formal V\&V aspects of acting.

\vspace{1em}

The paper is organized as follows. After introducing above the \emph{acting} decisional functionality and its skills, a state of the art of
the various approaches and formalisms, with respect to V\&V, is discussed in the next section concluding 
with a justification of the proposed approach and its originality.  Section~\ref{sec:proskill} presents the \ps{}\footnote{\textbf{Pro}\emph{cedural} \textbf{Skill}.} acting language, its
primitives and its link to PRS.  The \fiacre{} formal framework we use is presented in Section~\ref{sec:formalfw}.  Section~\ref{sec:proskillformal} presents
how \ps{} primitives are mapped in \fiacre{} and then the type of formal properties one can show online but also at runtime. In
Section~\ref{sec:uav}, we introduce a real robotic example on which we successfully deployed our approach. A discussion in
Section~\ref{sec:conclusion} reassesses the pros and cons of the \proskill{}/\fiacre{} formal acting language, leading to possible future
work and then to the conclusion of the paper.

\section{Acting: state of the art and proposed approaches}
\label{sec:soa}

Over the years, the \emph{planning} field has managed to define and agree on a number of common languages and formalisms:
PDDL~\cite{Ghallab:1998vl}; NDDL~\cite{Frank:2003wxa}; ANML~\cite{Smith:2008tk}; HDDL~\cite{Holler:2020aa}; chronicle~\cite{Ghallab:1996wi};
etc.  Similarly, for robotic functional components (nodes), ROS~\citep{Quigley:2009tg} has emerged as a de facto standard, at least to
enable data and code sharing among roboticists.  In both domains, many would argue that it is good for the field as it allows systems
interoperability, comparison, competition, sharing, etc. while others would regret that it restricts and constrains the development of new
original approaches, as researchers may remain too focused on improving the associated algorithms.  Meanwhile, the
\emph{acting} field remains mostly organized along one system --- one language tandems.  Yet, we will see in the following sections
that there are large categories of approaches to model and execute skills to perform \emph{acting}, and we will stress how these approaches
may support V\&V.

\subsection{Reactive planners and procedural languages}

\emph{Acting} must cope with the real-world contingencies, although it cannot really plan, it can reuse some plans parts, or choose among a
set of ``reactive plans'' (e.g. RAP~\cite{Firby:1987tq}), and be able to synchronize and perform task decomposition (e.g.,
TDL~\cite{Simmons:1998wf}). These principles led to the development of many systems. Among them, PRS (Procedural Reasoning System) first
implementations (late eighties) were mostly used for procedural maintenance~\cite{Georgeff:1989vo}. It is only later that at LAAS, PRS
skills (called Operational Procedures) were used as an acting component in the LAAS architecture~\cite{Alami:1998wq,Ingrand:2007ug}. The
Propice-plan~\cite{Despouys:1999va} fork is an attempt to infuse some planning in the acting functionality of the robots. Overall, PRS was
deployed in many robotic experiments.  From a V\&V point of view, there are some attempts to transform PRS skills in Colored Petri
Net~\cite{De-Araujo:2004aa} and later to give a formal semantics to the PRS skill language~\cite{deSilva:2018vd}. But both studies remain
theoretical and were never applied, nor deployed to verify PRS skills. Similarly, some work was conducted to show how some formal properties on
TDL~\cite{Simmons:1998wf} skills could be proven using NuSMV~\cite{Simmons:2000vg}. But again, this study remained mostly theoretical.

More recently, PLP~\cite{Brafman:2016vn} proposes an acting language which borrows some of the PDDL constructs and provides means to monitor
the behavior of functional components. Later in \cite{Kovalchuk:2021vc}, the authors introduce a stochastic extension to their model based on
UPPAAL-SMC~\cite{David:2015kz}. With respect to V\&V, the goal to build a monitor of the controller is interesting, but it did not go as far
as deploying a formal V\&V.

Behavior Trees~\cite{Colledanchise:2018ub} are becoming quite popular in the robotic community. Initially developed for the video game
industry, their simple yet powerful execution tree mechanism (skills) based on \emph{sequence}, \emph{fallback} and \emph{parallel} nodes
has been quite successful in deploying reactive modular acting systems. Yet, they lack an explicit time representation and even if the authors propose
some mechanisms to check efficiency, safety, and robustness, those mostly rely on ad hoc procedures, not a fully equivalent formal model (but we
will see in the discussion (Section~\ref{sec:discussion}) that this could be seriously considered).

RAE ~\cite{Ghallab:2016uo} introduces the basic algorithms which can drive a procedural reactive acting engine. Inspired by the
PRS~\cite{Ingrand:2007ug} engine, it mostly differs on its semantics of task (RAE) vs. goal (PRS) and on its handling of event-based skills
execution. The later implementations of RAE/UPOM~\cite{Patra:2021aa} introduce a look ahead planning mechanism to evaluate future outcomes and help make better choices
while acting.

\subsection{State machines and programming language}

Finite state machines are often proposed as acting models. Yet, one should consider whose states are modeled? The states of the world and its
components, or the states of the acting program execution? In the former, they capture the acceptable states and state transitions of the
components, while in the later, they capture the possible execution states which can be more difficult to maintain (some argue that they are
then just a bunch of \code{GoTo} instructions).

Yet, there are acting systems which mostly rely on them. SMACH~\cite{Bohren:2010hi} is an acting system deployed along ROS\footnote{SMACH
  was supported in ROS1, but as of today, not in ROS2.} which models
skills with hierarchical state machines. The states in SMACH are execution states, each with the possible outcomes and one can have
multiple state machines active at once. Similarly, RAFCON~\cite{Brunner:2016kw} also proposes hierarchical state machines to program
robotics systems and allow the creation of concurrent flow controls. rFSM statecharts~\cite{Klotzbucher:2012wx} stresses the coordination
aspect of the acting system and proposes to use Harel statechart to implement it.

Interestingly, RMPL~\cite{Ingham:2001uga,Williams:2003ue} mixes hierarchical finite states machines, to represent the state of some devices
used in the experiment and, a programming language inspired by Esterel~\cite{Boussinot:1991wz} to program the skills which need to be
deployed and executed. This is an original combination, and we will further discuss it in Section~\ref{sec:discussion}.  More recently,
Proteus~\cite{McClelland:2021ur} also relies on similar hierarchical finite states machines for components but proposes a more classical
programming language to deploy the system.

\subsection{Acting/planning framework}

We must also consider systems using skills which are common (or share a significant part) between the planning (action model) and the
acting language and could benefit from their action model ``part'' to improve the verifiability of their skills. Indeed, most planners are
performing some kind of model verification, using states exploration, constraints satisfaction, etc. 

Cypress (i.e., SIPE/PRS) proposes the Act formalism~\cite{Wilkins:1995uw} to unify both functionalities, but this common skill/action model
remains mostly syntactic, and both engines pick the part they need from the Act representation.

The IDEA~\cite{Finzi:2004tj} and T-ReX~\cite{McGann:2008kz} also propose to merge the planning and acting representation, arguing that they
are similar processes, with just a different horizon and response time. Organizing the state variables of the problem along different
planners/reactors, they use constraints to specify the acceptable state variable values and value transitions. But writing the proper
constraints to perform acting and planning ends up being quite tedious and error prone.

OMPAS~\cite{Turi:2023aa,Turi:2024aa} proposes an acting language whose skills can be automatically transformed in temporal chronicles to be
used by a temporal planner for some limited horizon planning and help the acting component to make better informed choices. This is an
interesting approach as it bridges skill models and action models.

\subsection{Programming skills within a formal framework }
\label{sec:psdwaff}

Last, there are also many \emph{acting} approaches which explicitly rely on some well-founded formal frameworks:
\begin{description}

\item[Situation calculus]  Many systems propose Situation Calculus as an underlying model for planning and acting:
  GOLEX~\cite{Hahnel:1998tl}, YAGI~\cite{Eckstein:2020tf}, Golog++~\cite{Matare:2021uc} to name a few. Yet despite the formal underlying
  framework, their deployment in real systems remains confidential.

\item[Petri net] is a well-known formalism to model concurrent systems and is supported by many formal tools. It is used in many acting
  systems: PROCOSA~\cite{Barbier:2006aa}, ASPIC~\cite{Lesire:2018tw}, Petri net~\cite{Costelha:2012du} and Hierarchical Petri
  Nets~\cite{Figat:2022we}, etc. In~\cite{Albore:2023aa}, the authors propose a skill robot language (RL) and \cite{Pelletier:2023aa}
  SkiNet, a verification tool which automatically maps RL in Time Transition Systems (Time Petri Net with data). This approach has some
  commonalities with the one we propose\footnote{We will see that we borrow some of the RL semantics in \ps{}.}, and we discuss their
  difference in Section~\ref{sec:discussion}.

\item[Synchronous language] Many approaches rely on a synchronous language ``hypothesis'' (communication and computation in no
  time). Historically MAESTRO/ORCCAD~\cite{CosteManiere:1992bz,Espiau:1996vu,Kim:2005gi}, relying on Esterel, already nailed down the idea
  of an acting skill language able to map in a formal model.  ReX~\cite{Kaelbling:1988ab}/GAPPS~\cite{Kaelbling:1988aa} can also be seen as
  a synchronous approach skill implementation.  Plexil~\cite{Jonsson:2006tl} executes skills modeled with different types of nodes along
  some control structures. A formal, but adhoc, extension is presented in~\cite{Dowek:2007vd}.

\item[Robot Chart] The Robot Chart~\cite{Cavalcanti:2017gh} framework is also deeply grounded in formal models, and numerous extensions have
  been proposed. But the language does not seem to grab much popularity or use in the robotics community which probably discards it as too
  complex or cumbersome to use.

\end{description}

\subsection{Our approach}
\label{sec:ourapproach}

On one hand, most of the approaches presented in Section~\ref{sec:soa} (Subsection~\ref{sec:psdwaff} apart) lack support for V\&V with an
automatic and systematic translation of their skills in a formal framework. On the other hand, the ones presented in
section~\ref{sec:psdwaff} have a strong potential to provide formal V\&V, yet they often remain difficult or cumbersome to use for
roboticists as most of them require some knowledge of the underlying formalism. Moreover, none of them provide a model which can be used
both for offline \emph{and} runtime verification.

In the survey~\citep{Bjorner:2014ta}, the  authors write:
\begin{quote}
  \normalsize{\emph{``We will argue that we are moving towards a point of singularity, where specification and
      programming will be done within the same language and verification tooling framework. This will help break down
      the barrier for programmers to write specifications.''}}
\end{quote}

Similarly, in~\citep{Nordmann:2016tj} the authors survey robotic DSL, and they argue that:
\begin{quote}
  \normalsize{\emph{``Both communities should foster collaboration in order to make formal methods more practicable and
      accepted in robotic software development and to make DS(M)L approaches more well-founded in theory to foster work
      in the field of model validation and verification.''}}
\end{quote}

Following these advices again (we have already presented a formal framework for functional components~\cite{Dal-Zilio:2023aa}), we now want to
extend the use of formal models toward the \emph{acting} component. Hence, the proposed approach presents the following characteristics:
\begin{itemize}
\item Programming the acting component using our proposed skill language does not require deep knowledge in formal verification models and
  languages. Some understanding of PDDL and PRS like languages is sufficient.
\item The language can be automatically, fully, and unequivocally translated in \fiacre{}~\cite{Berthomieu:2008vo,Berthomieu:2020vo}, a
  preexisting formal language,  with a clear semantics,
\item The obtained \fiacre{} formal model can be used both offline with model checking to prove some properties of the \emph{acting}
  program, but also online to execute the formal model on the robot.
\item The skill programs operational semantics is defined by the translation to the formal specification in \fiacre{} and is supported by
  the execution of the formal model itself on the robot.
\end{itemize}

As we will see, providing a skill language, whose translated formal model is directly executing on the robot, has several  advantages: it increases the
credibility that the skills are doing what they are intended too, and it improves the acceptability by roboticist as no extra step is
needed to run the skills within a formal framework, so there is no need to develop a specific ``skill'' execution engine.
\label{clar3}

Note that we have focussed our state of the art on acting frameworks deployed for autonomous robots, and we do not cover programming
languages used for example for industrial robots on assembly lines, often proprietary of the robot manufacturer.
\label{clar8}

\section{The \ProSkill{} language}
\label{sec:proskill}
\label{clar2}

The \ps{} language and specifications rely on four basic primitives: state variables, events, basic skills,
and composite skills. The first three are inspired by similar objects in RL~\cite{Albore:2023aa}.

For each of these objects, we present their operational semantics, i.e., their role, and how they relate to each other, the platform, and the environment. 

\subsection{State variables}

State variables are variables which take their value in either a bounded natural number or among several enumerated values. They usually
represent variables whose values are going to be tested and set in the various skills, to reflect the ``current'' state of the robot and the
environment. They are defined with statements such as the ones on Listing~\ref{lst:psv}.  They always have an initial value
(lines~\ref{lst:psv}.\ref{ll:init1} and~\ref{lst:psv}.\ref{ll:init2})\footnote{Listing lines are referenced with the $<$listing
  number$>$.$<$line number$>$, example:~\ref{lst:psv}.\ref{ll:init1}, Listing~\ref{lst:psv}, line:~\ref{ll:init1}.}, and for the latter, one
can allow all transitions to any other values, or more restrictively to some limited number of values
(line~\ref{lst:psv}.\ref{ll:transitions}).

\begin{lstlisting}[label={lst:psv},caption={Example of State Variables definition},columns=fullflexible, numbers=left, xleftmargin=15pt, language=ProSkill,float=!ht]
(defsv flight_levels
  :init 1  (*@ \label{ll:init1} @*) 
  :min 1
  :max 3)

(defsv battery
  :states (Good Low Critical) ; the possible values the state variable can take
  :init Good ; the initial value  (*@ \label{ll:init2} @*) 
  :transitions ; use the :all keyword if all transitions are allowed,  (*@ \label{ll:transitions} @*) 
    ((Good Low)(Low Critical) ; list them otherwise.
     (Critical Low)(Low Good)))
\end{lstlisting}

In this example the state variable \sv{battery} can change from any values to any other values except from \svv{Critical} to
\svv{Good} and vice versa.

\subsection{Events}
\label{sec:psevent}

Events correspond to uncontrollable external events which can occur at any time. The effects of an event are to produce some \emph{state
  variables} value change.

In the following example, some robot sensors will issue the \event{battery\_to\_critical} event, and this will update the \sv{battery}
state variable (see Listing~\ref{lst:pevt}).

\begin{lstlisting}[label={lst:pevt},caption={Example of an \event{event} definition}, numbers=left, xleftmargin=15pt, columns=fullflexible,language=ProSkill,float=!ht]
(defevent battery_to_critical
  :effects (battery Critical))
\end{lstlisting}

Note that if an event leads to a forbidden state variable transition, an error is raised.

\subsection{Basic skills}
\label{sec:basicskill}

The basic skills are the lowest level skills. They are the ones which act on the underlying robotic system, i.e., they launch robot commands
with arguments, and when the command completes, the basic skill sets the status (\code{no\_status}, \code{success}, \code{failure}, \code{failed\_inv} or
\code{interrupted}) and the results, if any. The underlying commands can be programmed in ROS~\citep{Quigley:2009tg} (e.g., using ROS
Actions), or other robotic frameworks, such as \GenoM{}~\cite{Dal-Zilio:2023aa}.

Although the syntax is slightly different, the basic skills semantics is equivalent to the RL one described in~\citep[Section 4.5, Figure
7]{Albore:2023aa}. Listing~\ref{lst:pbs} shows an example, withdrawn from the drone use case we present in section~\ref{sec:uav}, of a basic
skill specifying and commanding the \skill{takeoff} of the drone.

\begin{lstlisting}[label={lst:pbs},caption={Example of a Basic Skill \skill{takeoff}}, numbers=left, xleftmargin=15pt, columns=fullflexible,language=ProSkill,float=!ht]
(defskill takeoff
  :input ($height float $duration float)      ; The expected arguments
  :precondition (not_moving (motion Free)     ; a list of taged  (state_variable value)
           on_ground (flight_status OnGround) ; all four preconditions need to be satisfied
           battery_good (battery Good)
           origin_valid (localization_status Valid))
  :start (motion Controlled)                  ; state_variable to set upon starting this skill
  :invariant (in_control (:guard (motion Controlled)) ; all the guards are monitored while
              battery (:guard (~ (battery Critical))  ; the drone takeoff, if one
                        :effects (motion Free)))      ; fails, its effects, if any, are set 
                                                    ; and the action is interrupted
  :time_interval [1,3]   ; this is the interval of time in seconds this action takes  (*@ \label{ll:ti} @*) 
  :action (takeoff)      ; this is linked to the C/C++ code which will be invoked  (*@ \label{ll:action} @*) 
  :interrupt (:effects (motion Free)) ; when interrupted, set this state_variable
  :success at_altitude                           ; success, and their effects on state_variable
          (:effects (motion Free)                ; postcondition are just checked for
           :postcondition (flight_status InAir)) ; model consistency  (*@ \label{ll:pc1} @*) 
  :failure (grounded (:effects (motion Free)     ; they are not enforced
             :postcondition (flight_status OnGround))  (*@ \label{ll:pc2} @*) 
            emergency (:effects (motion Free)
             :postcondition (flight_status InAirUnsafe))))  (*@ \label{ll:pc3} @*) 
\end{lstlisting}

Most of the basic skill fields are self-explanatory, and their semantics is the expected one. State variable values are checked (in \field{precondition}, \field{invariant},
\field{postcondition}) and set (in \field{start} and \field{effects}). Note that all \field{guard} and \field{effects} are enforced, but
\field{postcondition} (lines~\ref{lst:pbs}.\ref{ll:pc1}, \ref{lst:pbs}.\ref{ll:pc2}
and~\ref{lst:pbs}.\ref{ll:pc3}) are just checked (they are expected as an effect of the success or failure, but not
enforced). \field{time\_interval} (line~\ref{lst:pbs}.\ref{ll:ti}) is a specification of the duration the action takes, and the
\field{action} field (line~\ref{lst:pbs}.\ref{ll:action}) specifies the command to call on the robot when this skill runs.

From an execution point of view, when a basic skill is called, its status is first set to \code{no\_status}. Its \field{preconditions} are
checked, if they are not satisfied, the skill returns. If they are, the \field{start} effects are set and the \field{action} is called. The
\field{invariant} is continuously checked while the \field{action} command executes. If the \field{invariant} fails, the skills returns with
\code{fail\_inv} status. If the skill is interrupted, the \field{interrupt} \field{effects} are set, the \field{action} command is
interrupted and the skill returns the \code{interrupted} status. The execution time is checked against the \field{time\_interval} values, and an error
is raised if it gets out of bound. When the \field{action} command completes (with success or failure), the
proper \field{effects} are set, \field{postconditions} are checked, and the returned status is set to \code{success} or \code{failure},
accordingly.

\subsection{Composite skills}
\label{sec:compskill}

Composite skills are hierarchically above basic skills and act more like a programming language where one can call other skills (basic or
composite), test the returned status and values, test state variable values, branch according to these tests, loop, wait
for some time or conditions, and even execute multiple branches in parallel. Composite skills are strongly inspired by PRS/Propice-plan
procedures~\citep{Ingrand:1996uj,Despouys:1999va}.

Listing~\ref{lst:pcs} shows an example of a composite skill withdrawn from the experiment presented in section~\ref{sec:uav}.  The fields
common to the basic skills have the same semantics and are treated alike. The most important and new field here is the \field{body} field
(line~\ref{lst:pcs}.\ref{ll:body}), instead of the \field{action} field (line~\ref{lst:pbs}.\ref{ll:action}) found in basic skills, which lists the program
instructions to execute when the skill is called.

\begin{lstlisting}[label={lst:pcs},caption={Example of a Composite Skill: \skill{uav\_mission}},columns=fullflexible,language=ProSkill,numbers=left,xleftmargin=15pt,float=!ht]
(defskill uav_mission
  :time_interval [60 , 120] ; expected min and max time to perform this skill.
  :success mission_accomplished ; success and 
    (:effects (mission_status Succeeded)) ; its effect
  :failure mission_failed ; failure and 
    (:effects (mission_status Failed)) ; its effect
  :start (mission_status Ongoing) ; setting the mission status SV upon starting
  :body (*@ \label{ll:body} @*) 
    ((start_drone) ; This will call the start_drone basic skill   (*@ \label{ll:startd} @*) 
     (^ (localization_status Valid)) ; wait for the localization to be Valid   (*@ \label{ll:wait1} @*) 
     (takeoff height 3.0 duration 0) ; call the takeoff basic skill (3 meters)  (*@ \label{ll:takeoff} @*) 
     (if (= takeoff.status success) ; if the takeoff status is success  (*@ \label{ll:success1} @*) 
        (// ((camera_survey)) ; execute two branches in parallel camera_survey on one  (*@ \label{ll:parallel} @*) 
            ((goto_waypoint x 1 y 2 z 3 yaw 0 duration 0) ; navigation on the other one
             (^ 2)     ; wait 2 seconds   (*@ \label{ll:wait2} @*) 
             (goto_waypoint x -3 y -2 z 4 yaw 1.4 duration 0) ; navigating to a new position
             (camera_survey.interrupt))) ; interrupt the camera_survey skill  (*@ \label{ll:interrupt} @*) 
     (if (= goto_waypoint.status success) ;the last goto_waypoint was a success   (*@ \label{ll:success2} @*) 
        (landing) ; call the landing skill
        (if (= landing.status success) ; if successful   (*@ \label{ll:success3} @*) 
           (shutdown_drone) ; call the shutdown_drone skill
           (printf "Mission Accomplished") ; print a message
           (success mission_accomplished)))) ; return the mission_accomplished success
    (printf "Mission failed") ; otherwise, print and report failure.
    (failure mission_failed)))
\end{lstlisting}

Note that for any \skill{skill}, one can access \code{skill.status} (lines~\ref{lst:pcs}.\ref{ll:success1}, \ref{lst:pcs}.\ref{ll:success2}
and~\ref{lst:pcs}.\ref{ll:success3}) and \code{skill.res} to respectively check the status (\code{no\_status}, \code{success},\code{failure},
\code{failed\_inv} or \code{interrupted}) and the result of the \skill{skill}'s last call and execution. Moreover, one can also call
\code{skill.interrupt} to interrupt an executing \skill{skill} (line~\ref{lst:pcs}.\ref{ll:interrupt}).

Instructions with \code{(\textasciicircum ...)} instruct the execution to wait for the \sv{localization\_status} to be \svv{Valid}
(line~\ref{lst:pcs}.\ref{ll:wait1}), or 2 seconds (line~\ref{lst:pcs}.\ref{ll:wait2}).

The call to a skill set its status and the programmer must proceed accordingly. For example, the succes of \skill{takeoff} basic skill
(line~\ref{lst:pcs}.\ref{ll:takeoff}) need to be checked explicitly (line~\ref{lst:pcs}.\ref{ll:success1}) to proceed accordingly.

\subsubsection*{Monitor skills}
\label{sec:monitorskill}

Monitor skills are a particular type of composite skill which are called immediately upon starting the execution of the \ps{} program (like
the main composite skill). Their first instruction is usually to wait for a condition they monitor, then they execute the rest of the body like any other
composite skill.

\begin{lstlisting}[label={lst:pmon},caption={Example of a Monitor Skill to gently land the drone when the \sv{battery} level
becomes \svv{critical}.},columns=fullflexible, numbers=left, xleftmargin=15pt, language=ProSkill,float=!ht]
(defskill monitor_battery_critical
  :monitor t ; Mark this skill as a monitoring one (hence it gets started upon startup)
  :body ((^ (battery Critical)) ; wait until the battery becomes critical
         (printf "We will try to land the drone safely...") ; print a message
         (set_velocity vx 0.0 vy 0.0 vz -0.1) ; force landing by setting -10cm/s vertical speed
        ))
\end{lstlisting}

\section{A formal framework for offline and runtime verification: \fiacre{}, language, models, and tools}
\label{sec:formalfw}

Although the goal of this paper is not to present in detail the formal framework we use, it is necessary to clarify some terminology and
give some explanations to make it self-contained. The more curious readers can check the specific papers and websites referenced
below.

\subsection{Terminology, models, languages, and tools}

We first clarify the following terms:

\begin{description}

\item[Time Petri nets] \cite{Berthomieu:1991wv} are an extension from regular \emph{Petri nets} model where each transition has a time
  interval (by default $[0,\infty)$) which specifies that the transition is sensitized and can be fired only during this time interval.

\item[TTS] Time Transition Systems are an extension of \emph{Time Petri nets} with data, and where transitions can call data processing
  functions.

\item[\tina{}] stands for ``TIme petri Net Analyzer'', it is a toolbox for the editing, simulating and analysis of \emph{Petri nets}, \emph{Time
    Petri nets} and \emph{TTS}. Among these tools, \tool{sift} and \tool{selt} can be used to respectively build the set of
  reachable states of the model and check LTL properties on the model.\footnote{\url{https://projects.laas.fr/tina/index.php}}

\item[\fiacre{}] stands (in french) for "Intermediate Format for Embedded Distributed Component Architectures''.  \fiacre{} is a formally
  defined language to compositionally represent the behavioral and timing aspects of embedded and distributed systems for formal
  verification and simulation purposes. \fiacre{} formal specifications can be compiled in a formally equivalent \emph{TTS} with the
  \texttt{frac} compiler.\footnote{\url{https://projects.laas.fr/fiacre/index.php}}

\item[\hfiacre{}] is an extension of the \fiacre{} language to make the specified model ``executable'' by adding \emph{Event Ports} and
  \emph{Tasks} both linked to C/C++ functions.
  
\item[\hippo{}] is an engine to execute \emph{TTS} obtained from \hfiacre{}
  specifications~\citep{Hladik:2021vt}.\footnote{\url{https://projects.laas.fr/hippo/index.php}}

\end{description}

This framework has been deployed in numerous projects and applications,\footnote{\url{https://projects.laas.fr/fiacre/papers.php}} and, not
surprisingly, is also the framework we used to validate and verify the functional components of our robotics
experiments~\cite{Dal-Zilio:2023aa}.

\subsection{\fiacre{} semantics}

\begin{figure*}[!ht]
\begin{center}
\includegraphics[width=0.97\textwidth]{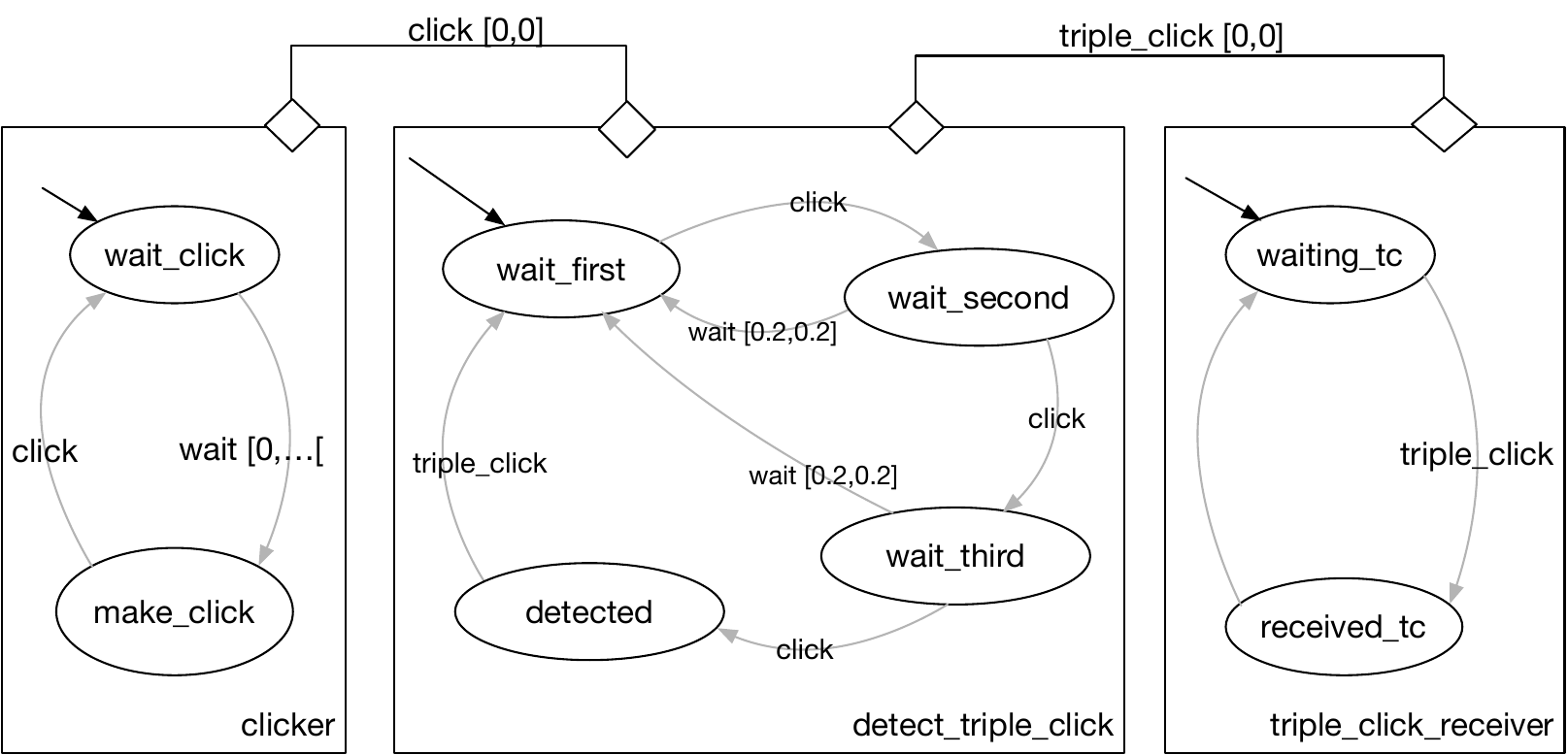}
\caption{The \fiacre{} processes modeling the \fiacre{} specification on Listing~\ref{lst:ftct}.}
\label{fig:tcd-tina}
\end{center}
\end{figure*}

Although we refer the reader to specific papers and websites (see above) for the formal model presentation and the tools, we think it is
important to get an idea of the semantics of the \fiacre{} language with a small example: the specification of a mouse triple clicks
detector.  This example is specified on Listing~\ref{lst:ftct} along the illustration Figure~\ref{fig:tcd-tina}. It defines three \fiacre{}
processes, each with its own automata. The first process, \process{clicker}, produces a \event{click} at any time. It waits between $0$ and
$\infty$ and then synchronizes on the \port{click} port with the \process{detect\_triple\_click} process. This second process has four
states, waiting for synchronization on \port{click}, or that the maximum acceptable time between two clicks (\qty{0.2}{\sec}) has
elapsed. Note the \code{select} on the \state{wait\_second} and \state{wait\_third} states, which is a non-deterministic choice which will
be explored by the model checker. When the \state{detected} state is reached, a synchronization on the \port{triple\_click} is made and this
allows the transition of the \process{triple\_click\_receiver} process to the \state{received\_tc} state.

Following these process specifications, one component is specified by putting three process instances in parallel
(line~\ref{lst:ftct}.\ref{ll:par}) and connecting them with two ports (line~\ref{lst:ftct}.\ref{ll:ports}). This example is intentionally
simple, but the \fiacre{} language supports complex data type, directional ports exchanging data (in and out), local and global variables, tests,
switch/case, guards on transitions and calls to functions (internal to \fiacre{} or external with C/C++ code) making complex computations. More
complex \fiacre{} specifications will be introduced and can be found in \ref{app:fbsh} and \ref{app:fcsh}.

\begin{lstlisting}[caption={\fiacre{} specification for a triple click detector (\fiacre{} offline version).}, numbers=left, xleftmargin=15pt, label={lst:ftct}, language=fiacre]
process clicker [click:sync] is // synthesize clicks and sync them on its port at any time
states wait_click, make_click

from wait_click
   wait [0, ...[; // wait any time from zero to infinity
   to make_click

from make_click
   click; // issue a click sync on the Fiacre port
   to wait_click

process detect_triple_click [click:sync,triple_click:sync] is
states wait_first, wait_second, wait_third, detected

from wait_first
   click;         // first click  (*@ \label{ll:click1} @*) 
   to wait_second

from wait_second
   select        // we wait either
     wait [0.2,0.2]; //  exactly 0.2 second
     to wait_first // then reset the detector
   []
     click;       // or for the second click   (*@ \label{ll:click2} @*) 
     to wait_third // whichever comes first
   end

from wait_third
   select        // again for the third click
     wait [0.2,0.2];
     to wait_first
   []
     click;       // third  (*@ \label{ll:click3} @*) 
     to detected
   end

from detected
   triple_click; // sync on the triple_click port
   to wait_first

process triple_click_receiver[triple_click:sync] is
states waiting_tc, received_tc

from waiting_tc
   triple_click; // just wait for a sync on this port
   to received_tc

from received_tc
   /* do what needs to be done when a TC has been detected */
   to waiting_tc

component comp_tc is //we now specifiy the component

port click:sync in [0,0], triple_click:sync in [0,0] // two ports  (*@ \label{ll:ports} @*) 

par * in // 3 processes composed in parallel  (*@ \label{ll:par} @*) 
   detect_triple_click[click, triple_click]
|| clicker[click]   
|| triple_click_receiver[triple_click]
end

comp_tc // this instantiates the component

// some properties to check
property ddlf is deadlockfree  // deadlock free (TRUE) (*@ \label{ll:ddf} @*) 
prove ddlf

property cannot_receveice_tc is absent comp_tc/3/state received_tc (*@ \label{ll:rtc} @*) 
prove cannot_receveice_tc // we cannot detect a triple click (FALSE)
\end{lstlisting}

\subsection{Offline formal verification}
\label{clar6}

The \fiacre{} language and the \tina{} toolbox used here mostly rely on LTL (SE-LTL state/event LTL~\cite{Chaki:2004aa} to be more
accurate). LTL already offers a rich language and a lot of flexibility to prove some logical and temporal properties on the model. By
default, the \tool{sift} tool computes the reachable states of the system, and one can either check properties on the fly (reachability of a
state), or show the absence of a state (but for this, the whole reachable state has to be build). Moreover, even if \fiacre{} and \tina{} do
not provide per se a timed logic (e.g. TCTL or MITL), the toolbox provides some patterns~\cite{Abid:2014aa} which allow the verification of
properties with explicit time (e.g. to prove that at most x units of time will elapse between these two states). Nevertheless, there are
more powerful frameworks than LTL (which examines single temporal trace), such as CTL, which adds quantifiers over traces and branching-time
constructs to standard LTL. In fact the \fiacre{}/\tina{} framework also allows (at some cost) to check CTL formula.

The \tool{frac} compiler compiles the \fiacre{} specifications on Listing~\ref{lst:ftct} in an equivalent TTS. Then using the \tool{sift}
tool from the \tina{} toolbox, one builds the set of reachable states of the system, and then using \tool{selt}, one checks the properties
included in the \fiacre{} initial specifications as well as new properties, if needed. There are many other tools available in the \tina{}
toolbox, which the interested reader can check.

Listing~\ref{lst:ftct} proposes some properties which will be checked on this specification: is the model deadlock free (line~\ref{lst:ftct}.\ref{ll:ddf})?
which ends up being TRUE. Can it succeed in detecting a triple click (line~\ref{lst:ftct}.\ref{ll:rtc})? (by checking that reaching the \state{received\_tc}
is impossible), and this is FALSE, so it means that the model can detect a triple click. More complex properties such as proving that there
is at most \qty{0.4}{\sec} between the first and the last click could be added, etc. 
 
Note that the verification approach used by the \tina{} tools is based on model checking, and as such suffers from state
explosion~\cite{Clarke:2012uv} which can jeopardize the usefulness of such an endeavor. Nevertheless, we will see in the results section of
the example we present in section~\ref{sec:results}, that we are still able to produce interesting nontrivial results.


\subsection{\hfiacre{} runtime extensions}

Although the \fiacre{} language was initially designed for offline verification, it has been extended with two primitives which enable it to
be used for runtime verification~\citep{Hladik:2021vt}, by connecting the model to C/C++ functions which send events or execute some
commands. To distinguish it from pure \fiacre{} we call the extended version \hfiacre{}.

The goal of the \hfiacre{} runtime version is to make the model ``executable'' while being connected to the real world.

\begin{figure*}[!ht]
\begin{center}
\includegraphics[width=0.97\textwidth]{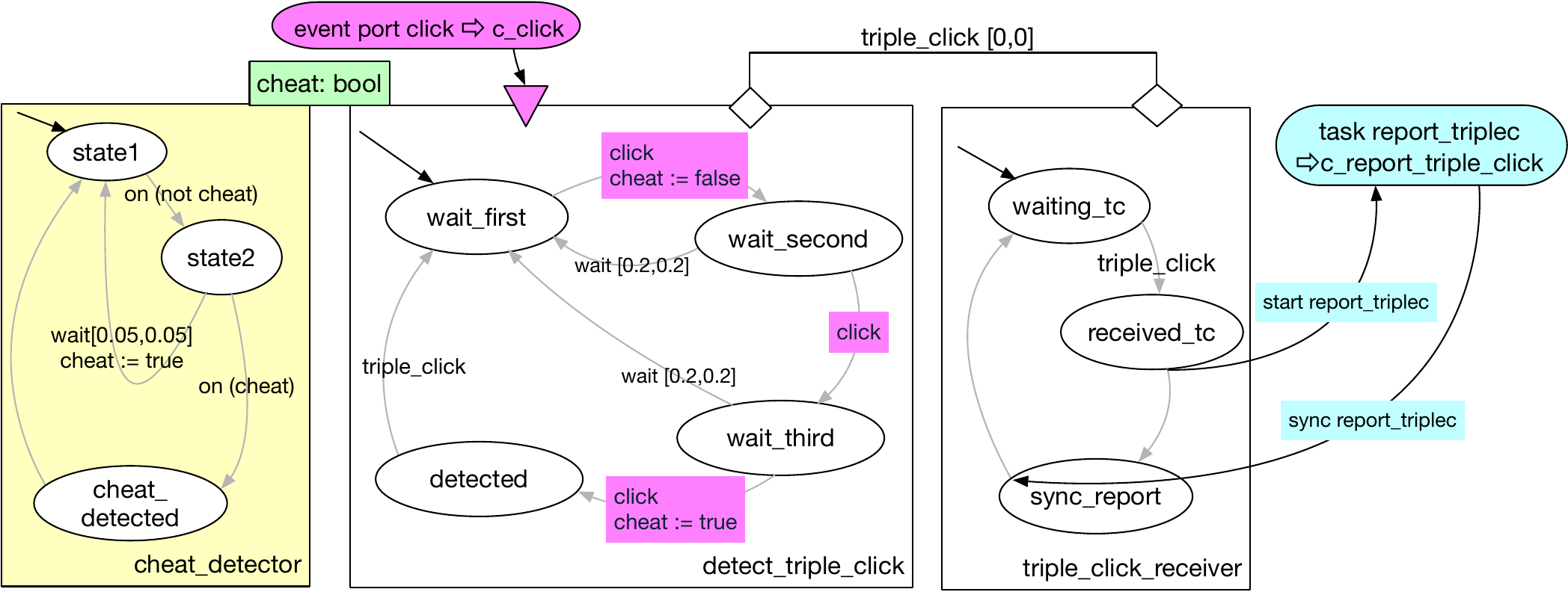}
\caption{Illustration of the \hfiacre{} program on Listing~\ref{lst:ftch}.}
\label{fig:tcd-hippo}
\end{center}
\end{figure*}

Listing~\ref{lst:ftch} (along Figure~\ref{fig:tcd-hippo}) shows the executable version of the specification on Listing~\ref{lst:ftct}.

\begin{description}
\item[Event ports] are declared in the preamble of the specification (see line~\ref{lst:ftch}.\ref{ll:event_port}) and they associate a C
  function to a \fiacre{} port. In this example, the event \event{click} is associated with the \code{c\_click} C/C++ function. Whenever
  this port is among the possible transitions (lines~\ref{lst:ftct}.\ref{ll:click1},~\ref{lst:ftct}.\ref{ll:click2}
  and~\ref{lst:ftct}.\ref{ll:click3}), the C/C++ function is called, and the port is activated when the function returns (the C/C++
  functions can take and return \fiacre{} typed arguments).
\item[Tasks] are also declared in the preamble (see line~\ref{lst:ftch}.\ref{ll:task}) and they associate a task (here
  \task{report\_triplec}) to a C/C++ function (here \code{c\_report\_triple\_click}), which will be called asynchronously upon a
  \code{start} (see line~\ref{lst:ftch}.\ref{ll:start}) and will enable the corresponding \code{sync} (see
  line~\ref{lst:ftch}.\ref{ll:sync}) when the C/C++ function returns. Here also, values can be passed upon calling the task and returned
  when complete.
\end{description}

\begin{lstlisting}[caption={\hfiacre{} processes implementing a triple click detector.}, numbers=left, xleftmargin=15pt, label={lst:ftch}, language=fiacre]
event click : sync is c_click // declare the Fiacre event port which transmits click (*@ \label{ll:event_port} @*) 
task report_triplec () : nat is c_report_triple_click // The C/C++ function this task calls  (*@ \label{ll:task} @*) 

process detect_triple_click [triple_click:sync] is 
// this process is exactly the same than in the regular Fiacre version
// only the click port is now an event port

process triple_click_receiver[triple_click:sync] is
states waiting_tc, received_tc, sync_report
var ignore : nat

from waiting_tc
   triple_click;
   to received_tc

from received_tc // show an example of an external call
   start report_triplec();  (*@ \label{ll:start} @*) 
   to sync_report

from sync_report
   sync report_triplec ignore;  // wait until the call return (*@ \label{ll:sync} @*) 
   to waiting_tc

component comp_tc is
port triple_click:sync

par * in
   detect_triple_click[triple_click]
|| triple_click_receiver[triple_click]
end

comp_tc
\end{lstlisting}

In this example, we have replaced the \process{clicker} process, which synced \port{click} at any time, by the
\event{click} event port (in purple), and we have added a task (\task{report\_triplec} in light blue) to execute when we synchronize with a
\event{triple\_click} in the \process{triple\_click\_receiver} process. The rest of the model remains the same, so we moved from a
model to specify a triple click detector, to a program/controller which implements it. The specification is now also a program.

\subsection{Runtime (online) verification}
\label{sec:rv}

The resulting \hfiacre{} model is then compiled (with \tool{frac}) in TTS and linked with the \hippo{} engine and the C/C++ functions needed to
run the model (i.e.,  \code{c\_click} and \code{c\_report\_triple\_click}).  The \hippo{} engine literally runs the resulting TTS model
and makes the appropriate calls (in separate threads) to the C/C++ functions associated with \emph{event ports} and \emph{tasks}. Note that we can
also enrich the model with properties to check on the fly by adding a monitor process. For example, if this controller is used in a video
game where the game programmer wants to detect sequence of triple click produced so fast that they are not humanly possible, and must come
from a cheating device, then we could add a new \process{cheat\_detector} process (on the left Figure~\ref{fig:tcd-hippo} and Listing~\ref{lst:fcd}) with three
states, sharing a Boolean variable \code{cheat} set to false in the transition to \state{waiting\_second} of process
\process{detect\_triple\_click}, and true in the transition to \state{detected}.

\begin{lstlisting}[caption={\process{cheat\_detector} process detecting a cheating device by monitoring the \code{cheat} Boolean variable.}, numbers=left, xleftmargin=15pt, label={lst:fcd}, language=fiacre]
process cheat_detector(&cheat:bool) is

states state1, state2, cheat_detected

from state1
  on (not cheat); // guard on (not cheat)
  to state2
  
from state2 // cheat was set to false
  select // either 
     wait [0.05,0.05]; // 50 ms elapsed
     cheat := true; // reset the cheat variable
     to state1 // go back to monitoring
  []
     on (cheat); // cheat became true again before the 50ms above.
     to cheat_detected //caught cheating
  end

from cheat_detected
  // the player is cheating, do what needs to be done.
  to state1
\end{lstlisting}

In the \process{cheat\_detector} process \state{state1}, it would \emph{guard} on \code{(not cheat)} and then would transition to \state{state2}
and wait either \qty{50}{\ms} or \code{(cheat)}. If \code{cheat} becomes true before the \qty{50}{\ms} (i.e., a super-fast triple
click has been issued), then it transitions to \state{cheat\_detected} and reports a suspicious behavior, otherwise, it sets \code{cheat} to
true and goes back to \state{state1}.

So, we synthesize a controller which runs the specification. This is one of the critical advantages of the \fiacre{} framework: the same
formal model can be checked offline and run online. Beyond verification, on one hand, you get the real controller performing what the model
specifies, and on the other hand, observing the behavior of the running model confirms that your initial specifications do what you intended
it to do, and thus the properties you check offline are indeed, applied to the same online ``behaviorally'' good model.  Of course,
observing the proper behavior of the running specified model is a necessary but not sufficient condition. Yet, it is better than having no
link between the specification and the execution.  Note that if the runtime model is wrong, you can start debugging it (the same way you
debug regular programs). Also note that from a formal point of view, for a given model, all the traces of the \hfiacre{} version are
included in the ones of the \fiacre{} version.

\section{The \ps{} language and its mapping in \fiacre{} models}
\label{sec:proskillformal}

We introduced the \ps{} language primitives in Section~\ref{sec:proskill}, we now show how each of them translates to \fiacre{} variables
and processes, how they interact (through \fiacre{} ports) and how we instantiate and compose them in a global \fiacre{} component.

\begin{figure*}[!ht]
\begin{center}
\includegraphics[width=0.97\textwidth]{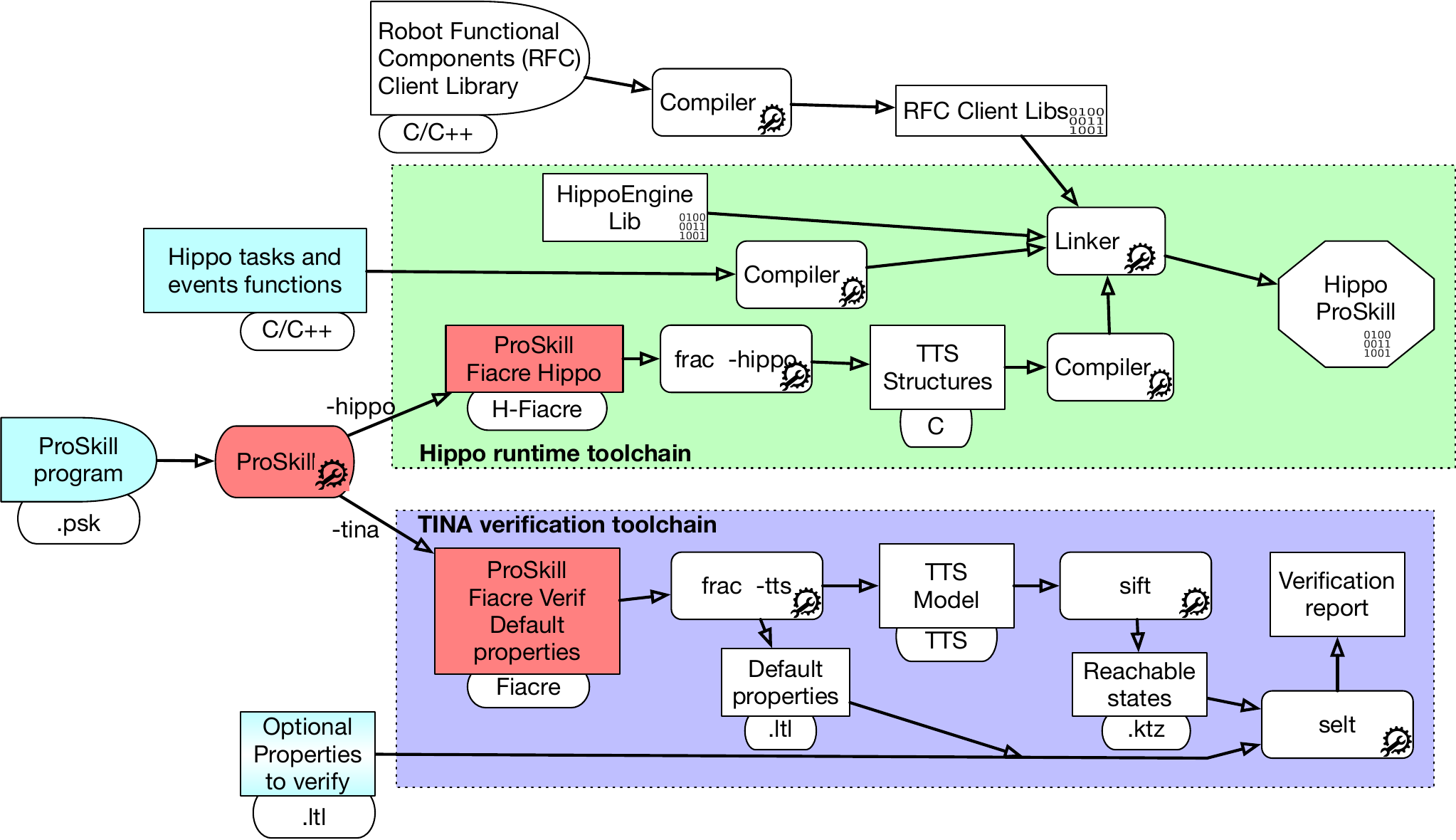}
\caption{The \ps{} to \fiacre{}/\hippo{}-\tina{} workflow. Only the data in the blue boxes need to be provided by the programmer. \ps{} (in
  light red) synthesizes the two models, the rest is fully synthesized.  In light green, the workflow for the \hippo{} runtime verification
  version, and in light purple, the workflow for the \tina{} offline verification version.}
\label{fig:toolchain_fiacre}
\end{center}
\end{figure*}

Before getting into the detail of the produced models, we presents on Figure~\ref{fig:toolchain_fiacre} the overall workflow from the \ps{}
program to the executable version (top part in green), and the verifiable version and its analyzed properties report (bottom part in
purple).  One should keep in mind, that the robotic programmers just write the \ps{} program, the C/C++ code which glue basic skill actions
and \ps{} events to the real robot and the additional LTL properties to check (in blue), the rest is fully synthesized and compiled
automatically.
\label{clar1}

\subsection{State variables \fiacre{} version}

\ps{} state variables are automatically transformed to \fiacre{} variables which can only take values among the one specified in the \ps{}
specifications. Their allowed value transitions are automatically specified in a \fiacre{} process.  Note that if
the state value change is not consistent with the \ps{} specified ones, an error occurs (lines~\ref{lst:fsv}.\ref{ll:error1},
\ref{lst:fsv}.\ref{ll:error2}). For natural number state variables, we must specify the accepted values interval.  Listing~\ref{lst:fsv}
shows how we implement
the state variables specified in Listing~\ref{lst:psv}.

\begin{lstlisting}[caption={\fiacre{} type and \fiacre{} process specifying the State Variable acceptable value changes.}, numbers=left, xleftmargin=15pt, label={lst:fsv}, language=fiacre]

type sv_flight_levels is 1..3 /* state variable integer types*/

process sv_battery_automata (&battery:sv_battery) is
    // for each sate variable, an automata enforces the allowed transition
states Good, Low, Critical, error

from Good
    wait [0,0];
    select
        on (battery = Low);
        to Low
    []
        on (battery = Critical);
        to error // forbidden transition  (*@ \label{ll:error1} @*) 
    end

from Low
    wait [0,0];
    select
        on (battery = Good);
        to Good
    []
        on (battery = Critical);
        to Critical
    end

from Critical
    wait [0,0];
    select
        on (battery = Good);
        to error  // forbidden transition (*@ \label{ll:error2} @*) 
    []
        on (battery = Low);
        to Low
    end
\end{lstlisting}

\subsection{Events \fiacre{} version}

The example introduced in Section~\ref{sec:psevent} is automatically translated to the following simple \fiacre{} process. Each \ps{} event
become a \fiacre{} sync port which will either be connected to an \emph{Environment} \fiacre{} process in the offline verification; or to
the real environment, with event ports\footnote{Although \ps{} events end up mapped in \fiacre{} event ports in the \hfiacre{} version, they
  should not be confused.}, in the runtime version (i.e., some C/C++ functions will trigger and synthesize the event accordingly).

\begin{lstlisting}[caption={\fiacre{} process specifying how the \event{battery\_to\_critical} event is handled.}, numbers=left, xleftmargin=15pt, label={lst:fevt}, language=fiacre]
process event_battery_to_critical_automata
        [battery_to_critical : sync] // a Fiacre sync port on which we get the event
        (&battery: sv_battery) is // the battery state variable

states start_

from start_
    battery_to_critical; // when this port interact
    battery := Critical; // the battery value is updated to Critical
to start_
\end{lstlisting}

\subsection{Basic skills \fiacre{} version}
\label{sec:bsfv}

\begin{figure*}[!ht]
\begin{center}
\includegraphics[width=0.97\textwidth]{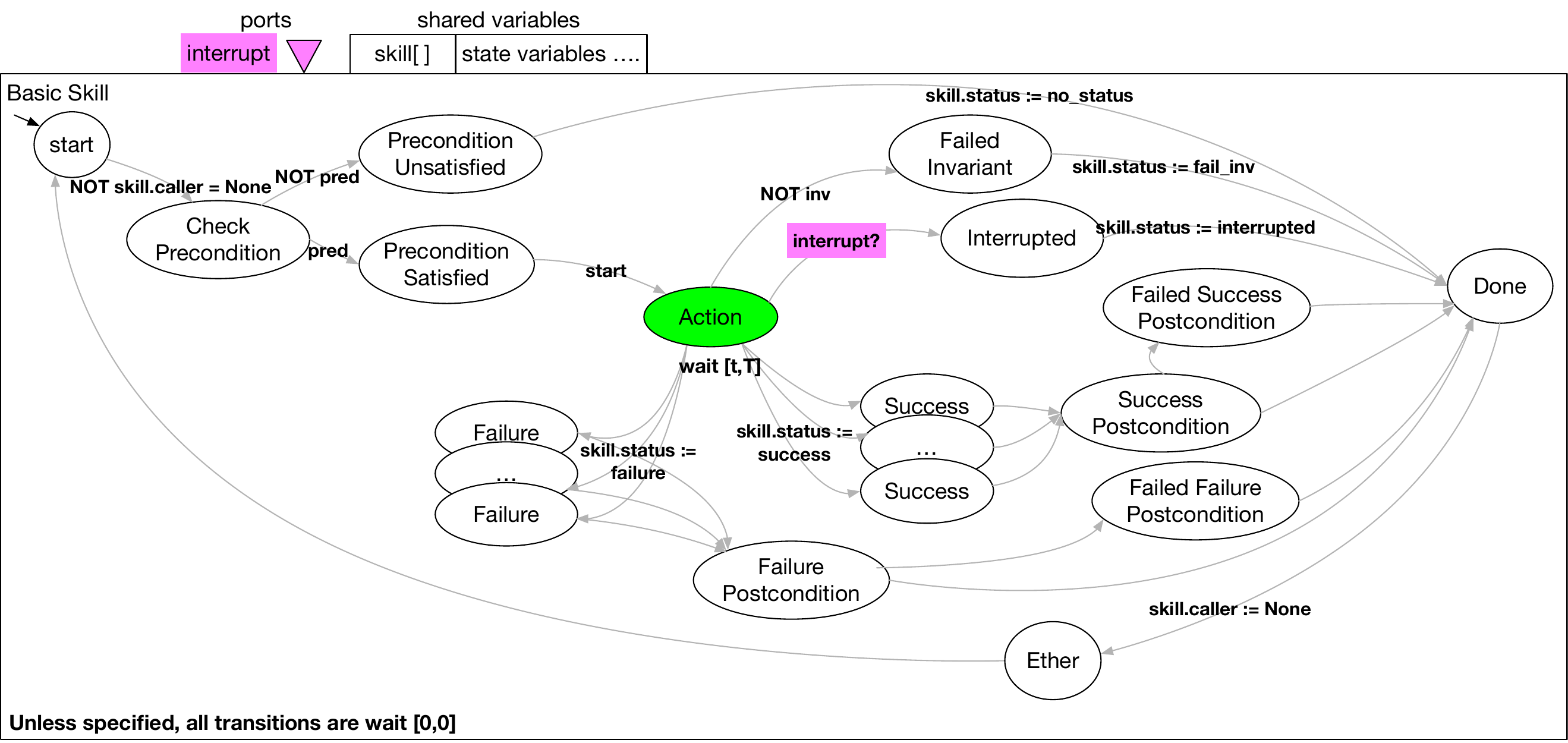}
\caption{The Offline Verification Basic Skill Automata \fiacre{} model (\tina{}).}
\label{fig:skill-tina}
\end{center}
\end{figure*}

\begin{figure*}[!ht]
\begin{center}
\includegraphics[width=0.97\textwidth]{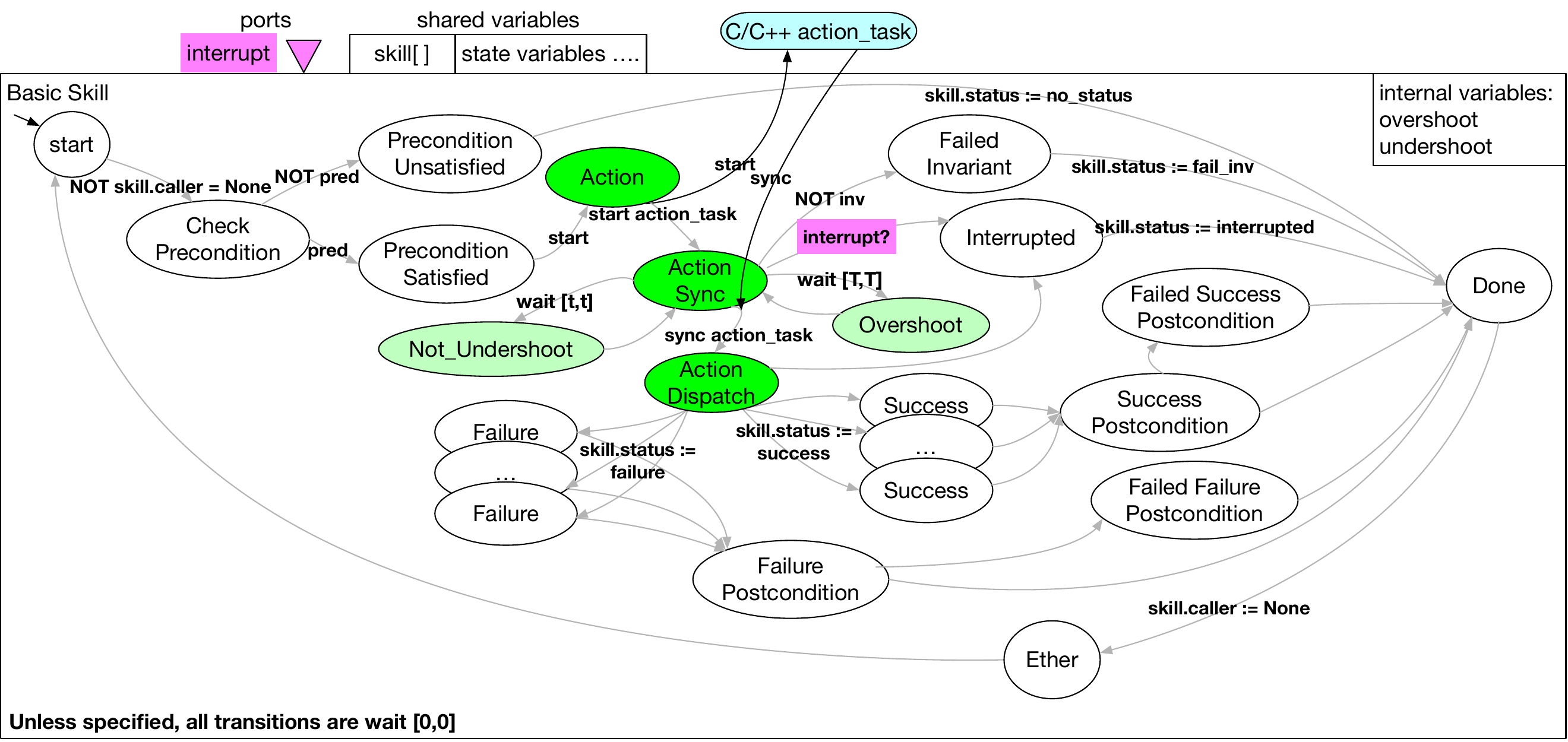}
\caption{The Online Verification Basic Skill Automata \hfiacre{} model (\hippo{}).}
\label{fig:skill-hippo}
\end{center}
\end{figure*}

As expected, the basic skills are the ones ``executing'' the commands on the robot.  The automata \fiacre{} offline version of the basic
skill presented Section~\ref{sec:basicskill} is illustrated on Figure~\ref{fig:skill-tina}, while the automata \hfiacre{} runtime is
illustrated on Figure~\ref{fig:skill-hippo} and listed in~\ref{app:fbsh}, Listing~\ref{lst:fbsh}.

An important \fiacre{} variable which appears in the various \fiacre{} listing is the \code{skill$[]$} record array, indexed for each skill (basic
and composite) and for each parallel branch, whose elements (a \fiacre{} record) contain for each skill/branch, information as whether it is
currently running or not, which skill called it, what are its argument, and what are its last status and result.

As expected, the only differences between these two models (\fiacre{} and \hfiacre{}) is how they handle the execution in the \state{Action}
state. In the offline version which will be explored by model checking, the \state{Action} state has a large \code{select}
(line~\ref{lst:fbsc}.\ref{ll:select}) which lists the various possible outcomes of the action execution:
\begin{itemize}
\item An external interruption from the \port{interrupt} port of the process,
\item An invariant guard failure,
\item All the possible successes and failures within the $[t,T]$ time interval specified to perform the action (e.g., line~\ref{lst:pbs}.\ref{ll:ti}).
\end{itemize}

In the runtime verification version, the \state{Action} state leads to the call to the \fiacre{} task \code{action\_task}, and then waits
(\code{sync}) in the
\state{Action Sync} state one of the possible outcomes:
\begin{itemize}
\item An external interruption from the \port{interrupt} port,
\item An invariant guard failure,
\item \state{Not\_undershoot} and  \state{Overshoot} states which will monitor in real-time these timing errors,
\item The return of the \task{Action} task call (with the \code{sync}) which makes a transition to the \state{Action Dispatch} state, which
  will dispatch to \state{Failure} or \state{Success} states according to the returned status.
\end{itemize}

Listing~\ref{lst:fbsc} shows a compact version of the offline \skill{takeoff} basic skill with only two states, \state{idle} and
\state{run}. Semantically, it is equivalent to the expanded version, but it is better suited for model checking as it removes some null time
state transitions interleaving in the model.  Note how a skill gets activated by guarding on its \code{caller} \code{skill} field not equal
to \code{None} (line~\ref{lst:fbsc}.\ref{ll:called}). Similarly, it gives control back to the caller by setting it back to \code{None}
(line~\ref{lst:fbsc}.\ref{ll:callret}). Similar \fiacre{} code can be found in Listing~\ref{lst:fbsh} in appendix~\ref{app:fbsh} which shows
the complete \hfiacre{} version of the \skill{takeoff} basic skill.

\begin{lstlisting}[caption={\fiacre{} process specifying the \skill{takeoff} basic (offline verification compact version) (Figure~\ref{fig:skill-tina-compact}).}, numbers=left, xleftmargin=15pt, label={lst:fbsc}, language=fiacre]
process skill_takeoff
    [interrupt_takeoff: sync] // interruptible skills get this port
    (&skill: skill_array, &flight_status: sv_flight_status, &target: sv_target, 
     &mission_status: sv_mission_status, &localization_status: sv_localization_status, 
     &motion: sv_motion, &battery: sv_battery, &camera: sv_camera) is

states idle, run

from idle
    wait [0,0];
    on (not (skill[takeoff].caller = None)); // a composite skill has called us  (*@ \label{ll:called} @*) 
    if (not (invariant_active(skill, flight_status, target, mission_status,
             localization_status, motion, battery, camera))) and // invariant not propagating
        ((motion = Free) and (flight_status = OnGround) and (battery = Good) and 
         (localization_status = Valid) and true) then // precondition satisfied
        motion := Controlled; // start state_variable set
        skill[takeoff].inv_active := true; // skill is active and its invariant monitored
        skill[takeoff].status := no_status; // reset status
        to run // go to the run state
    else // the precondition is not satisfied
        skill[takeoff].caller := None; // the call is not possible
        to idle
    end

from run // Action: (takeoff)
    select // the wait values correspond to the values specified in the time_interval field  (*@ \label{ll:select} @*) 
        wait [1, 3]; // at_altitude success 
        skill[takeoff].val := takeoff_ret_val(takeoff_success_at_altitude);
        motion := Free;   // success effect
        skill[takeoff].status := success // report success
    []
        wait [1, 3]; // grounded failure
        skill[takeoff].val := takeoff_ret_val(takeoff_failure_grounded);
        motion := Free;  // failure effect
        skill[takeoff].status := failure // report failure
    []
        wait [1, 3]; // emergency failure 
        skill[takeoff].val := takeoff_ret_val(takeoff_failure_emergency);
        motion := Free;  // failure effect
        skill[takeoff].status := failure // report failure
    []
        interrupt_takeoff; // the interrupt event port sync
        motion := Free; 
        skill[takeoff].status := interrupted  // report interruption
    []
        wait [0,0]; 
        on (not (motion = Controlled));  // invariant guard failed
        skill[takeoff].status := failed_inv  // report failed inv.
    []
        wait [0,0];
        on (not (not (battery = Critical))); // invariant guard failed
        motion := Free; // effects of the invariant failure
        skill[takeoff].status := failed_inv
    end;
    skill[takeoff].inv_active := false;
    skill[takeoff].caller := None; // return to idle state and inform the caller we are done  (*@ \label{ll:callret} @*) 
    to idle
\end{lstlisting}

\begin{figure}[!ht]
\begin{center}
\includegraphics[width=0.99\columnwidth]{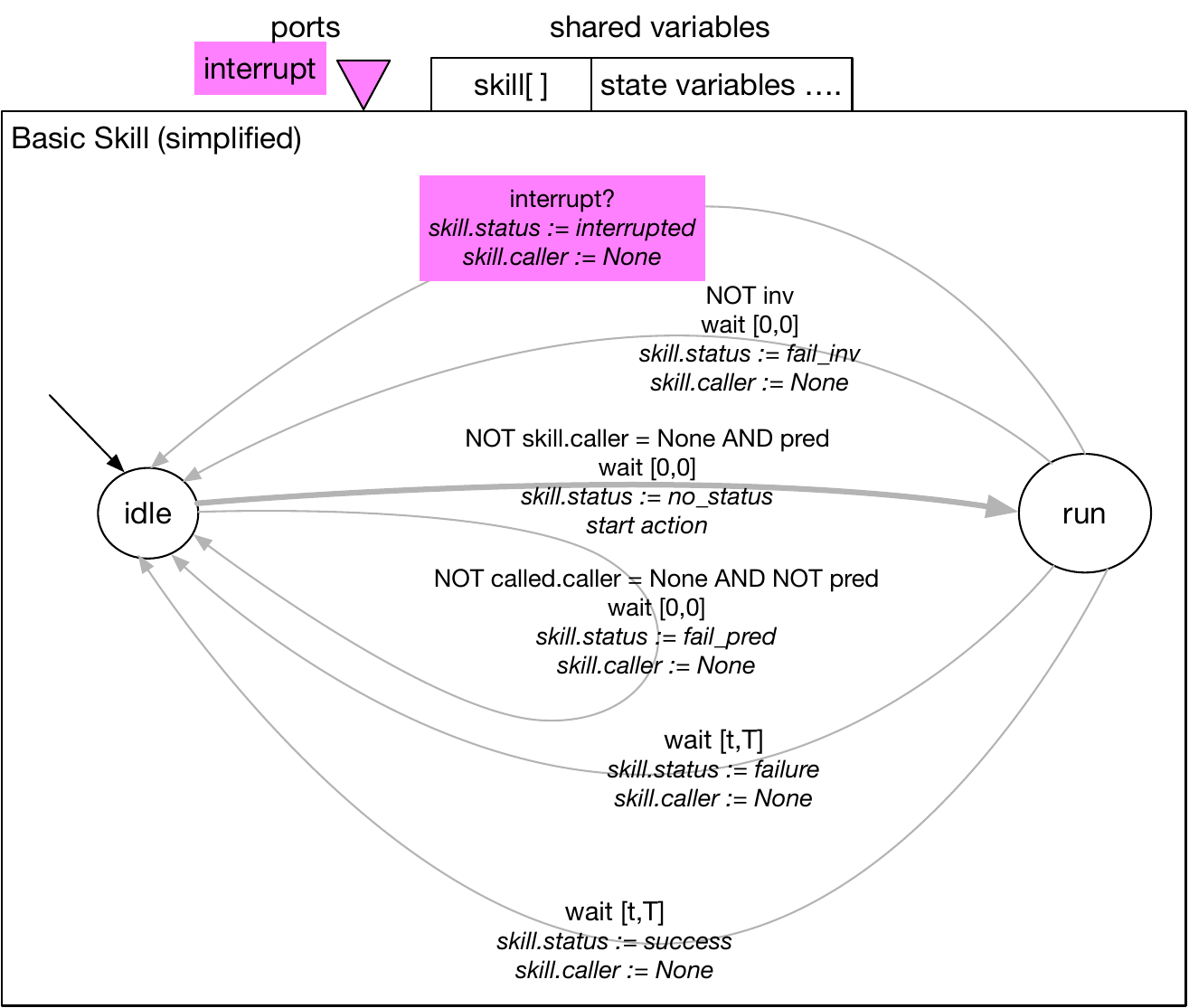}
\caption{A more compact version of the online verification basic skill \fiacre{} process (Listing~\ref{lst:fbsc}).}
\label{fig:skill-tina-compact}
\end{center}
\end{figure}

Figure~\ref{fig:skill-tina-compact} illustrates the compact version of this \fiacre{} \skill{takeoff} basic skill process.

\subsection{Composite skills \fiacre{} version}

The mapping of the composite skills to \fiacre{} is also systematic and automatic. It follows the ``program to automata'' algorithm used in
PRS~\citep{Ingrand:1996uj}.  We illustrate it with some examples withdrawn from the composite skill in Listing~\ref{lst:pcs}.

\paragraph{Skill call} The call to \skill{takeoff} on line~\ref{lst:pcs}.\ref{ll:takeoff} translates to:
\begin{lstlisting}[numbers=left, xleftmargin=15pt, label={lst:fcall}, language=fiacre]
from NS5
    wait [0,0];
    skill[takeoff].caller := uav_mission; // transfer execution to the takeoff skill (*@ \label{ll:call} @*) 
to NS5_NS4_sync

from NS5_NS4_sync
    wait [0,0];
    on (skill[takeoff].caller = None); // wait until execution returns from the takeoff skill(*@ \label{ll:return} @*) 
to NS4
\end{lstlisting}

We set the caller \code{field} of the \code{skill} array element for the \skill{takeoff} index to the caller (\skill{uav\_mission})
(line~\ref{ll:call}). Note that as explained in the previous section, this will wake up the \skill{takeoff} skill
(line~\ref{lst:fbsc}.\ref{ll:called}). Then we wait in the next state until \skill{takeoff} is done (line~\ref{lst:fbsc}.\ref{ll:callret})
which triggers the guard (line~\ref{ll:return}).

\paragraph{Sequence}

Sequences are straightforward, just progress to the next state in the \fiacre{} automata.

\paragraph{Test and branching}

The test on the \skill{takeoff} success status on line~\ref{lst:pcs}.\ref{ll:success1} translates to:

\begin{lstlisting}[numbers=left, xleftmargin=15pt, label={lst:ftest}, language=fiacre]
 from NS4
    wait [0,0];
    if (skill[takeoff].status = success) then
        to N3_T // jump to state N3_T in case the test is true
    else
        to NS8 // jump to NS8 otherwise
    end
\end{lstlisting}

\code{N3\_T} and \code{NS8} correspond to the \code{true} and \code{false} branches following the test. In this example, there is no \code{else}
branch in the skill, hence it jumps to the end of the \code{if}.

\paragraph{Parallel branches}

Parallel branches involve the synthesis of additional \fiacre{} processes (one for each branch). The initial executing skill or branch relinquishes
the control to the parallel branches and waits until they all terminate. Hence the code line~\ref{lst:pcs}.\ref{ll:parallel} get translated
too:

\begin{lstlisting}[numbers=left, xleftmargin=15pt, label={lst:fpar}, language=fiacre]
from N3_T // Pass the execution to the two parallel branches...
    wait [0,0];
    skill[uav_mission_branch_1_0].caller := uav_mission; // ... to the camera_survey branch(*@ \label{ll:pbc1} @*)
    skill[uav_mission_branch_1_1].caller := uav_mission; // ... to the navigation branch (*@ \label{ll:pbc2} @*)
to N3_T_NS6_sync

from N3_T_NS6_sync // wait the control back from all branches 
    wait [0,0];
    on ((skill[uav_mission_branch_1_0].caller = None) and //wait camera_survey is done and(*@ \label{ll:pbr1} @*)
        (skill[uav_mission_branch_1_1].caller = None));  // the navigation is done (*@ \label{ll:pbr2} @*)
to NS6
\end{lstlisting}

One can see that using the \code{skill$[]$} array element for the considered caller, the control can be passed from the main body to the
branches (lines~\ref{ll:pbc1} and~\ref{ll:pbc2}) and then wait for the control to come back to the main body when they are finished
(lines~\ref{ll:pbr1} and~\ref{ll:pbr2}).

\ref{app:fcsh} lists the four \hfiacre{} processes synthesized to model the \skill{uav\_mission} (\S~\ref{sec:compskill}) composite skill.
Listing~\ref{lst:fcsh} is the main process and handles the execution of the main part of the body, while Listing~\ref{lst:fcshp1} and
Listing~\ref{lst:fcshp2} handle respectively the two parallel branches (one doing the \skill{camera\_survey}, while the other one does the
navigation with the two \skill{goto\_waypoint}). Last, listing~\ref{lst:fcshp3} presents the process in charge of checking the temporal over
and under shooting of the skill.

\subsection{Environment and final component}
\label{sec:envcomp}

For the resulting formal model to be properly fully analyzed and run, one needs to connect it to the real environment (or a model of it).
For the offline verification version, we synthesize a \fiacre{} process (e.g., Listing~\ref{lst:fcp}) which produces, at any time, all the
\ps{} events and interrupts, both modeled with \fiacre{} ports, present in the model (like the \process{clicker} process in the
triple click detector example).

\begin{lstlisting}[caption={The \fiacre{} process modeling the environment, which synthesizes all the possible \ps{} events and interrupts of the UAV experiment.}, numbers=left, xleftmargin=15pt, label={lst:fcp}, language=fiacre]
process basic_environment_drone // ports to produce all possible events and interrupts
        [localization_status_to_invalid : sync, ..., flight_status_to_in_air_unsafe : sync, 
         battery_to_good : sync, battery_to_low : sync, battery_to_critical : sync, 
         interrupt_takeoff: sync,... ] is

states start_

from start_
    select
        localization_status_to_invalid
    [] 
        ... // ProSkill event ports sync removed for conciseness
        flight_status_to_in_air_unsafe
    [] 
        battery_to_good
    [] 
        battery_to_low
    [] 
        battery_to_critical
    [] 
        interrupt_takeoff
    [] 
        ... // interrupt ports sync removed for conciseness

    end;
to start_
\end{lstlisting}

This is perfect for model checking and to explore all the possible occurrences of asynchronous events and interrupts in the model. Yet, if
the application environment follows a more restrictive pattern, then the user can modify this process which hopefully may lead to a smaller
(but no larger) reachable states set.

For the runtime version, we also provide a simple process handling the \fiacre{} ports in the model. But in this case, this environment process has two
\fiacre{} event ports, one for \ps{} events and one for interruptible skills (they can be interrupted by an external ``signal''). These two \fiacre{} event
ports are each linked to a C/C++ function which will appropriately trigger when these external events or interrupts are received on the real robot.

\subsection{The final \fiacre{} component}

We have presented all the processes which model the various \ps{} objects. To create a complete \fiacre{} component, these processes must be
instantiated, and their \fiacre{} ports properly connected. Thus, the final \fiacre{} component defines:

\begin{itemize}
\item The \fiacre{} variables definition including all the state variables used in the experiment and an array of a \code{skill} record/structure which stores
  the various information needed for each skill instance (section~\ref{sec:bsfv}).
\item The \fiacre{} ports definition for all the possible \ps{} events and interrupts.
\item The \fiacre{} process instances:
  \begin{itemize}
  \item State variables automata, which control the allowed transitions for enumerated state variables.
  \item Event effects, whose \fiacre{} event port will be connected to the Event and Interrupt environment processes (see Section~\ref{sec:envcomp}).
  \item Basic skills with an interrupt port, for each interruptible skill.
  \item Composite skills
  \item Composite skill temporal watchdog to monitor possible (reachable) overshoot or undershoot.
  \item Composite skill parallel branches, if any.
  \item Event and interrupt environment process for \ps{} event.
  \end{itemize}
\end{itemize}

\subsection{Offline and online formal verification of \ps{} programs}
\label{sec:ltlprop}

The \fiacre{} model of the \ps program can be used in both offline and online (runtime) verification.

\subsubsection{Offline formal verification}
\label{sec:ofv}

As shown on Figure~\ref{fig:toolchain_fiacre}, the offline verification is often done in two steps, one to synthesize the set of reachable states of
the model (with \tool{sift}), and one to check some properties in the model (with \tool{selt}).

When the \fiacre{} model for \proskill{} program is synthesized, a number of default properties are also synthesized and checked with \tool{selt}.

\begin{itemize}
\item Check for each state variable that there is no forbidden transition (these transitions lead to an \state{error} state, so we check that
  it is not reachable).
\item For each basic skill that: it can run;  it can succeed, for all successes;  it can fail, for all failures;  its
  invariant may fail; and  it can be interrupted (all of these correspond to a specific state in the \fiacre{} model, so we check that
  they are reachable or not).
\item For composite skill, we also add if it can undershoot or overshoot its time interval specification (again, in the model we have error
  states corresponding to undershoot and undershoot, so  we check that they are not reachable).
\end{itemize}

Beyond these default properties, one can check any property involving  \fiacre{} process states, ports, external events, state variable values and time. For example, one can check that if we get the event that \event{Battery} is
\svv{Critical} while flying, then we will issue a \skill{landing} command in less than 1 unit of time.
\label{clar7}

Note that when one proves safety properties (i.e., to show that a dangerous state cannot be reached), we attempt to show that it can be
reached, and if it can (i.e. the property is false), an execution trace is given, to help the programmer to identify the problem and,
hopefully, fix it.

\subsubsection{Online (or runtime) verification}

Unlike the offline verification which ``model checks'' the \ps{} program, the runtime verification literally runs the same model.
Thanks to the \hippo{} engine, the TTS model is run, and commands are called, events are
received, and the robot executes the mission it has been programmed to do in \ps{}.

The \hfiacre{} model already contains several tests which may lead to warnings and error messages at runtime:
\begin{itemize}
\item overshoot and undershoot skill execution,
\item state variable illegal value transitions,
\item command illegal returned value.
\end{itemize}

In the \ps{}  experiment, the \hippo{} engine runs at \qty{100}{\Hz}, which is sufficient for the type of program and temporal
constraints we handle.

Note that similarly to the \process{cheat\_detector} monitor presented in Section~\ref{sec:rv}, we can augment the \hfiacre{} model with
processes to monitor specific situations requiring actions.

\section{An example: an UAV controller}
\label{sec:uav}
\begin{figure*}[!ht]
\begin{center}
\includegraphics[width=0.97\textwidth]{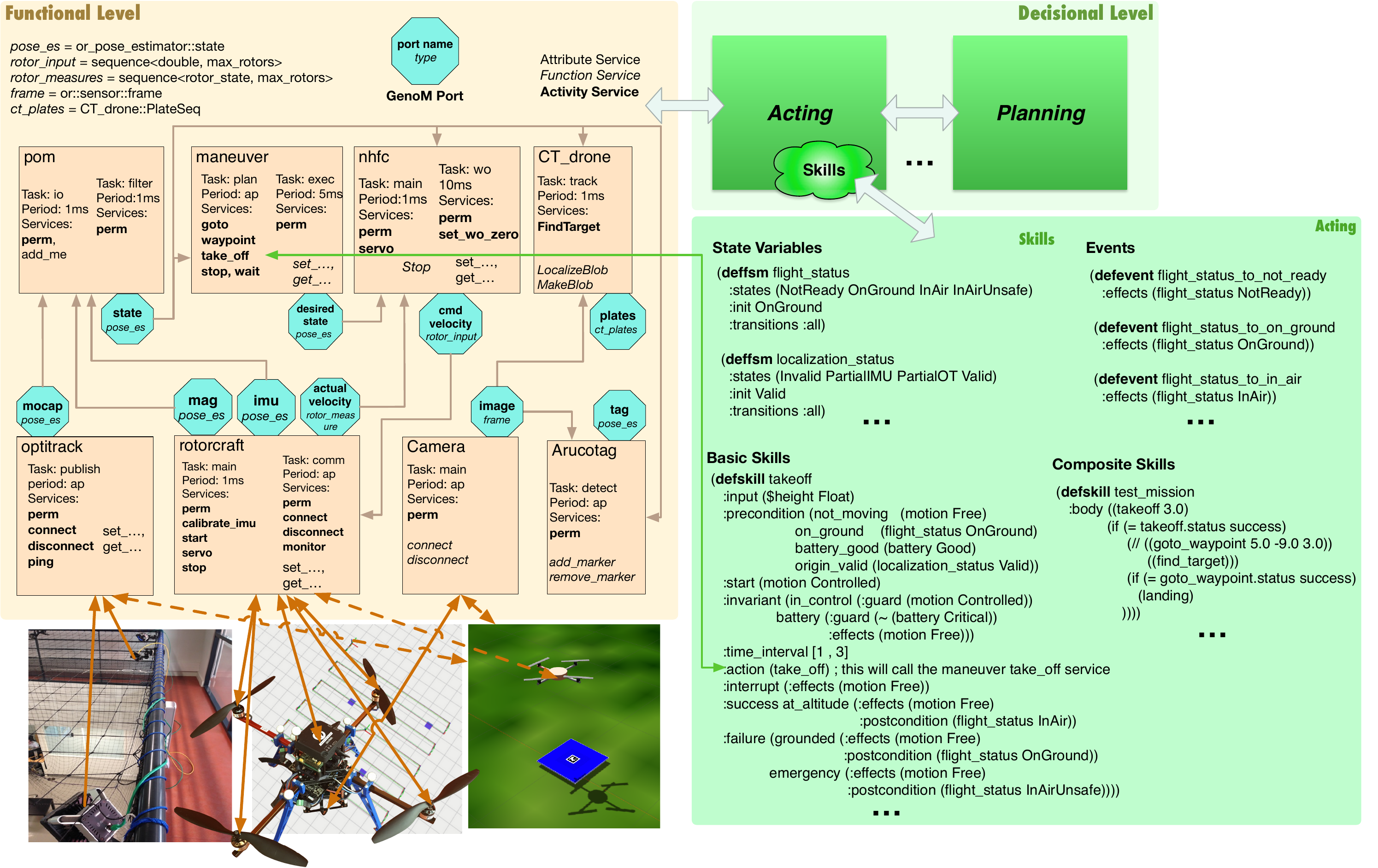}
\caption{Architecture of the drone experiment.}
\label{fig:archi-uav}
\end{center}
\end{figure*}

We demonstrate our approach with an UAV for which we program in \ps{} a survey mission. The functional layer of this experiment
(Figure~\ref{fig:archi-uav}) has already been presented in~\cite{Dal-Zilio:2023aa}, but suffice to say that it provides robust localization,
navigation, flight control and allows us to command the drone. It is deployed using the \GenoM{} specification language (which also maps in
a formal framework to validate and verify the functional components)~\cite{Dal-Zilio:2023aa}.

Although eight functional components are required for this experiment, the set of primitive commands available for the \emph{acting}
component are \skill{takeoff}, \skill{landing}, \skill{goto\_waypoint}, \skill{start\_drone}, \skill{shutdown\_drone}, and
\skill{camera\_survey}. Each of them has a corresponding basic skill (like the one presented on Listing~\ref{lst:pbs} for
\skill{takeoff}), to which we add a composite skill \skill{uav\_mission} (presented on Listing~\ref{lst:pcs}), and one skill to monitor for
the \sv{battery} level (Listing~\ref{lst:pmon}).

The state variables handled by the model are
\sv{flight\_status} with values: \svv{NotReady}, \svv{OnGround}, \svv{InAir}, \svv{InAirUnsafe} and \svv{Lost};
\sv{target}:  \svv{NotFound} and \svv{Found};
\sv{mission\_status}:  \svv{Unknown}, \svv{Ongoing}, \svv{Failed} and \svv{Succeeded};
\sv{localization\_status}:  \svv{Invalid}, \svv{Coarse} and \svv{Valid};
\sv{motion}:  \svv{Free} and \svv{Controlled};
\sv{battery}:  \svv{Good}, \svv{Low} and \svv{Critical};
and \sv{camera}:  \svv{On} and \svv{Off}.

Similarly, the available \ps{} external events are handling \sv{localization\_status} and \sv{battery\_status} updates.

The resulting \fiacre{} model has seven processes for the state variables, ten for external \ps{} events, six for basic skills, one for the
monitoring skill, one for a composite skill, along two for its parallel branches (Listings~\ref{lst:fcshp1} and~\ref{lst:fcshp2}) and one
for a watchdog for its time interval of the composite skill (Listings~\ref{lst:fcshp3}), and one for the environment process. This amount to
1400 lines of \fiacre{} code, which compiles in a TTS with 97 places and 153 transitions.\footnote{The resulting code can be found in the
  \code{examples} subdirectory of \url{https://redmine.laas.fr/projects/proskill/repository}.}

\subsection{Offline verification results}
\label{sec:results}

Building the set of reachable states of the complete model of the whole \ps{} program takes \qty{112}{\hour} \qty{17}{\minute} on
an AMD EPYC 7352 24-Core Processor (using one processor) and results in a set of reachable states which has 1 545 614 784 classes, 372 325
248 markings, 12 761 domains, 23 154 942 608 transitions.  We are able to check all the default properties defined in section~\ref{sec:ofv}.

Some results are puzzling, but correct. For example, model checking shows that the invariant on the \skill{landing} cannot fail. Indeed,
checking the code, its invariant cannot be falsified in this particular setup. Similarly, the main composite skill can
overshoot and undershoot. This can easily be explained as the wait for the \sv{localization\_status} to be \svv{valid} is unbounded, hence
it can take a priori any time.

All these results may indicate an error in the model, question some programming choices or confirm something expected. In any case, it
offers to the programmer a complete exploration of the execution possibility to study and analyze.

Of course, state explosion is the limiting factor here, and we will explore some possible improvements to address it in Section~\ref{sec:discussion}.

\subsection{Runtime verification results}
\label{sec:hresults}

We build a runtime version of the \ps{} specification and link it with the client libraries of the components which provide access to the
sensors (i.e., events) and commands of the drone: \modu{pom} for the \sv{localization\_status};  \modu{rotorcraft} for \sv{battery\_status}
but also the \skill{start\_drone} and \skill{shutdown\_drone} commands; \modu{camera} for \skill{camera\_survey} and
\modu{maneuver} for  \skill{takeoff}, \skill{landing} and \skill{goto\_waypoint}.  The \hippo{} engine executes the TTS model at
\qty{100}{\Hz}, it evaluates preconditions, invariants; tests conditions on state variables, or sets them when specified. It handles skill success
and failure and detects runtime overshoot or undershoot skills execution.

An important aspect of running the model itself is to confirm that the execution result is consistent with what the programmer has in
mind when he writes the \ps{} program. The \ps{} program becomes a formal model, and we run this model. We did several runs and showed
that the drone acts as expected, surveys the area, and lands gracefully when its \sv{battery} becomes \svv{Critical} (thanks to the
monitoring skill, Listing~\ref{lst:pmon}).

\vspace{1em}

Although we only illustrate our approach with one robotics platform/example, \ps{} can be used to implement the acting component of any
robots. The example code repository~\footnote{\url{https://redmine.laas.fr/projects/proskill/repository}} contains a number of \ps{}
programs written for other platforms.

\section{Limits, discussion, future work, and conclusion}
\label{sec:conclusion}

\subsection{Features and limits}
\label{sec:limits}

\ps{} offers many of the desired features for an \emph{acting} skill language. Skills can call each other with, if needed, a hierarchical
organization. Resources management (not described here) is easily handled with resource variables (a test for availability and reservation
is trivial to write in \fiacre{}, as all TTS transitions are atomically executed). Time management is provided, at the basic skill level,
but also in the composite ones. Commands, events, and interruptions are properly handled too. This is another strong point of the \ps{}
language: modeling nondeterminism (using \code{select}), raising from external events and from commands outcomes. Yet, the model
remains predictable as all the execution paths resulting from this nondeterminism will be explored by model checking. At the end, the
whole language can be mapped in a formal framework which opens a new realm for V\&V of robotic systems.

Still, \ps{} has not been defined with a formal semantics, but its operational semantics is defined by the semantics one gives to the
various instructions/specifications of the language. The main drawback of this approach in a verification process is that the model used for
the verification could end up with different semantic as the one programmer expect from the original language.  There is therefore a kind of
semantic gap between the verification and the implementation (this is the subject of~\cite{Hladik:2021vt}). Here this problem is somewhat
circumvented because the same model (obtained by translation from \ps{} to \fiacre{}) is used for offline verification \emph{and} for
runtime execution (i.e. in this case, the model \emph{is} the implementation). So the problem of proving that the semantics of \ps{} is
identical to that of \fiacre{} becomes somewhat irrelevant.  The semantics of \ps{} is provided by the transformation to \fiacre{} and the
equivalence between the verified model and the generated code is achieved by construction. One writes a program in \ps{}, translates it to
\fiacre{}, compiles and runs it. Thus we claim the two models are \emph{equivalent}, because even if we cannot prove formal equivalence
between the two implementations (\ps{} and \fiacre{}) the programmer can observe a ``correct'' behavior from the implementation running the
\fiacre{} model (like any programmer does when he or she observes the results of its program).

Yet there are some limits to the expressiveness and power of the \ps{} language. Some of them are due to our approach that the language must
map in \fiacre{}. For example, some basic types (e.g., float, string, etc.) are not currently handled by the \fiacre{} language. Therefore, this aspect
of the \ps{} language cannot be model checked, nevertheless, these variables values can be made available at runtime and can even be
involved in runtime verification.

Another manageable limitation is that the skills are not reentrant, and you cannot have skills with recursive calls (direct or indirect). In
short there can only be one instance of each skill active at one time. But nothing prevents the programmer from creating as many instances of a
particular skill as he may need.

If one considers most modern languages, those limits may seem extreme, but you have also to consider that often, plans given by
automatic planner are already fully instantiated, and most variables are bound to a value. Overall, this is a classical tradeoff between the
expressiveness of a language and its verifiability.

Uncertainty management is another area where \ps{} has little to offer. If we consider the \sv{localization} in our drone example, we
consider three discretized values \svv{Valid}, \svv{Coarse}, \svv{Invalid}. Currently the covariance of the position of the drone (computed
by the Kalman filter in \modu{pom}) is checked and the proper event is synthesized accordingly. Similarly, \fiacre{} does not provide
uncertainties on time intervals to model approximate time execution evaluation.  This choice is mostly justified by the goal of this work to
provide a formal model of skills and to run proof on this model. Of course, there are also formal model taking into account uncertainty. For
example ``Statistical Model Checking'' (e.g. UPPAAL SMC~\cite{Foughali:2019vj}), and we could consider extending \ps{} (and \fiacre{}) with
similar primitives, but these frameworks do not provide logical proof anymore (true or false), but statistical proof, by sampling the
reachable states space built using the probability distribution of the model.

The \ps{} language does not offer fancy algorithmic constructs which are sometimes provided to ease the programming task. We want to keep
the language as feature-limited as possible but complete enough to provide a powerful programming language. With the \code{if-then-else},
\code{while}, \code{do-until}, \code{goto}, \code{wait}, parallel branches execution and interrupt, we can write any of the complex programs
we encountered. But the language remains open to any new constructs, if they can either be programmed with the available ones or if they can be
translated to \fiacre{}.

\subsection{Discussion}
\label{sec:discussion}

The presented \ps{} framework has strong similarities with the SkiNet one~\cite{Albore:2023aa}, which also relies on the TTS formalism
and the \tina{} toolbox. Yet their modeling choices and their runtime verification approach differ:
\begin{itemize}
\item Unlike \ps{}, they directly synthesize TTS (i.e., they do not use the \fiacre{} language thus they skip the \texttt{frac} compilation
  step).
\item Even if TTS (an extension of Time Petri net) supports time, they do not currently model timing information in their basic and composite skills.
\item They do not synthesize a controller which executes the \ps{} program as we do, instead they synthesize a controller which will monitor the
  regular program.
\end{itemize}

These choices can be discussed and there are pros and cons to each approach, yet the differences are sufficient to justify separate
developments. Using \fiacre{} as an intermediate language has some strong advantages (legibility, expressivity, etc.), without any
performance impact. Moreover, executing the program with the \hippo{} engine greatly improves the confidence that the initial \ps{}
program does what the user wants, and providing we have a TTS player (\hippo{}), why not use it instead of writing another controller which
needs to be monitored using the TTS model?

Note that we designed the \ps{} language, having in mind the automatic transformation to \fiacre{}. But if we consider again the systems
presented in the state of the art in Section~\ref{sec:soa}, other languages/systems can probably be automatically and unequivocally
translated to \fiacre{}. Behavior trees and RMPL are probably suitable candidates for such mapping and this could lead to a valuable
``formally verified'' implementation~\cite{Ingrand:2024ab}.

Our previous work presented in~\cite{Dal-Zilio:2023aa} also produces a \fiacre{} model on which one can also perform offline and runtime
verification. Although we could consider merging the two models and then make some verification on the joint model, it is unlikely to be
effective considering we are already struggling to avoid state explosion in each of them. But for the runtime verification, this makes more sense, and we
have run experiments where all the functional components are run with one \hippo{} engine running at \qty{10}{\kHz} along another engine
executing the \emph{acting} specifications at \qty{100}{\Hz}.

Although the work presented here targets autonomous robots, we believe it could be of interest to program more classical industrial
robotic systems. Indeed, even the simplest automatic systems (e.g., lifts, driverless subways, assembly line robots, etc) can take advantage
of formal methods to certify their controller.

  \subsection{Future work}

Although the current version of the language is fully operational, there are several improvements to consider:
\begin{itemize}
\item Extend the data type handled directly by \fiacre{}.\footnote{The \fiacre{} developper are working on adding rational numbers and
    strings types.}
\item  To avoid state explosion while model checking, we could consider a more abstract version of the offline verification model, yet
  we need to keep a faithful and complete version of the model for the runtime to avoid a semantics gap between what the programmer intended,
  and what is executed.
\item Statistical Model Checking~\cite{Foughali:2019vj} could be used to better explore the possible execution branches resulting from the
  \fiacre{} \code{select} instructions, based on probability/distribution obtained from regular or simulated runs.
\item \ps{} basic skill specifications share some fields with \emph{planning} action models. So one could consider using these skill models in a
  planning process, as to produce plans which could be executed as composite skills. To some extent, this would relate to what is done in
  Propice-plan~\cite{Despouys:1999va} and in RAE/UPOM~\cite{Patra:2021aa}.
\item As we now have a formal model of the \emph{acting} component, and of the functional components of the robot~\cite{Dal-Zilio:2023aa},
  we could consider making one of the middleware connecting them.
\item Many autonomous robots are now deployed with humans present in the environment. One could extend our approach as to properly model the
  uncontrollable behaviors of human while they interact with the robots.
\end{itemize}

Nevertheless, all these improvements are extending the current implementation~\cite{Ingrand:2023aa}, while none requires a deep redesign of
the approach with respect to V\&V of autonomous robots.

\subsection{Conclusion}
\label{clar5}

\emph{Acting} is a critical decisional functionality of autonomous systems such as robots. We propose \ps{}, a new skill language to program
the acting component of robots. The basic building blocks of the \ps{} language are presented and so is the \fiacre{} formal framework. We
then show how the \ps{} primitives are mapped automatically and unambiguously in \fiacre{}. The obtained formal model can be used offline
with model checking to verify logical and temporal properties, but also online for runtime verification while executing the program/model.
The offline verification explores the model (for desirable or undesirable states, sequences, state variables values, etc.), while the
runtime verification runs it and enforces it.  We illustrate our integrated approach on a real platform: a drone executing a survey mission.
The fact that the formal model is the one executing at run time ensures both its operational “accuracy” (operational semantics) and enforce
the use of formal tools from specifications to verification and runtime execution. This is a major step toward deploying V\&V in the
autonomous robot \emph{acting} component and making it available to roboticists, even without any V\&V background. In an era where more and
more functional and decisional components are using deep learning based approaches, which are notably difficult to validate and verify, we
believe that deploying acting components which can act as monitoring and safety bag with respect to those learning based components is
essential to improve the trust we put in these robots.

\section*{Acknowledgement}
\noindent
We thank Bernard Berthomieu, Silvano Dal Zilio and Pierre-Emmanuel Hladik for their time, discussions and help while developing and
deploying the work presented here.

\ifARXIV

\else

\ifHAL
\bibliographystyle{abbrvnat}
\else
\bibliographystyle{elsarticle-harv}
\fi

\bibliography{master}
\fi

\newpage
\appendix
\section{Complete \hfiacre{} process for a basic skill}

Note: the \hfiacre{} code presented in these appendices is automatically synthesized from \ps{} specifications. As a result, some code may
seem cumbersome (e.g., \code{ignoreb} variables not used) or unnecessary (e.g., \code{if (true) then ... else ... end}, or \code{from x
  wait[0,0]; to y}). The \tool{frac} compiler keeps or simplifies these constructs as needed.

\label{app:fbsh}

\begin{lstlisting}[caption={The \hfiacre{} process specification of the \skill{takeoff} basic skill.}, numbers=left, xleftmargin=15pt, label={lst:fbsh}, language=fiacre]
process skill_takeoff
    [interrupt_takeoff: sync]
    (&skill: skill_array, &flight_status: sv_flight_status, &target: sv_target, 
     &mission_status: sv_mission_status, &localization_status: sv_localization_status, 
     &motion: sv_motion, &battery: sv_battery, &camera: sv_camera) is

states start_, check_precondition, precondition_satisfied, precondition_unsatisfied,
    action, action_sync, action_dispatch, error, action_sync_overshoot, 
    action_sync_not_undershoot, interrupted, failed_invariant, failure_grounded,
    check_failure_postcondition_grounded, failed_failure_postcondition_grounded, 
    failure_emergency, check_failure_postcondition_emergency, 
    failed_failure_postcondition_emergency, success_at_altitude, 
    check_success_postcondition_at_altitude, failed_success_postcondition_at_altitude, done, 
    ether

var ignoreb: bool, ret_val: takeoff_ret_val_type, overshoot, undershoot : bool

from start_
    wait [0,0];
    on (not (skill[takeoff].caller = None));
    skill[takeoff].status := no_status;
    // Skill 'takeoff' has been called
to check_precondition

from check_precondition
    wait [0,0];
    on (not (invariant_active(skill, flight_status, target, mission_status, localization_status, motion, battery, camera)));
    if ((motion = Free) and (flight_status = OnGround) and (battery = Good) and (localization_status = Valid) and true) then
        to precondition_satisfied
    else
        to precondition_unsatisfied
    end

from precondition_unsatisfied
    wait [0,0];
    // Skill 'takeoff' precondition ((motion = Free) and (flight_status = OnGround) and 
    // (battery = Good) and (localization_status = Valid) and true) UNsatisfied
to done

from precondition_satisfied
    wait [0,0];
    // Skill 'takeoff' precondition ((motion = Free) and (flight_status = OnGround) and 
    // (battery = Good) and (localization_status = Valid) and true) is satisfied
    motion := Controlled; 
    skill[takeoff].inv_active := true;
to action

from action // Action: (takeoff)
    // Skill 'takeoff' calling its action (start task)
    overshoot := false;
    undershoot := true;
    start Fiacre_takeoff_action_task (skill[takeoff]);
to action_sync

from action_sync
    select
        sync Fiacre_takeoff_action_task ret_val;
        // Skill 'takeoff' action returned (sync task)
        if undershoot then
            // WARNING: Action 'takeoff' undershoot 1 s (100 ticks)
            null
        end;
        to action_dispatch
    [] // undershoot
        on undershoot;
        wait [100, 100];
        to action_sync_not_undershoot
    [] // overshoot
        on not undershoot and not overshoot;
        wait [200, 200]; // 200 because we already waited 100 for not_undershoot
        // WARNING: Action 'takeoff' overshoot 3 s (300 ticks)
        to action_sync_overshoot
    [] //invariant
        wait [0,0];
        on (not (motion = Controlled));
        // Skill 'takeoff' failed invariant in_control: (motion = Controlled)
        skill[takeoff].val := takeoff_ret_val(takeoff_failed_inv_in_control);
        to failed_invariant
    [] //invariant
        wait [0,0];
        on (not (not (battery = Critical)));
        // Skill 'takeoff' failed invariant battery: (not (battery = Critical))
        motion := Free; 
        // Skill 'takeoff' failed invariant battery concluding effects: motion := Free; 
        skill[takeoff].val := takeoff_ret_val(takeoff_failed_inv_battery);
        to failed_invariant
    [] // we got an interrupt from the event port...
        interrupt_takeoff;
        motion := Free; 
        // Skill 'takeoff' interrupted from event port
        // Skill 'takeoff' is interrupting its own action takeoff
        ignoreb := Fiacre_takeoff_interrupt_action();
        to interrupted
    end

from action_sync_overshoot
    overshoot := true;
to action_sync

from action_sync_not_undershoot
    undershoot := false;
to action_sync

from action_dispatch
    wait [0,0];
    if (ret_val = takeoff_success_at_altitude) then
        to success_at_altitude
    end;
    if (ret_val = takeoff_failure_grounded) then
        to failure_grounded
    end;
    if (ret_val = takeoff_failure_emergency) then
        to failure_emergency
    end;
    if (ret_val = takeoff_interrupted) then
        skill[takeoff].val := takeoff_ret_val(takeoff_interrupted);
        // Skill 'takeoff' has been interrupted while executing its action.
        to interrupted
    end;
to error // a priori unreachable

from interrupted
    wait [0,0];
    skill[takeoff].inv_active := false;
    skill[takeoff].status := interrupted;
to done

from failed_invariant
    wait [0,0];
    skill[takeoff].inv_active := false;
    skill[takeoff].status := failed_inv;
to done

from success_at_altitude
    wait [0,0];
    skill[takeoff].val := takeoff_ret_val(takeoff_success_at_altitude);
    // Skill 'takeoff' success 'at_altitude'
    skill[takeoff].inv_active := false;
    motion := Free; 
to check_success_postcondition_at_altitude

from check_success_postcondition_at_altitude
    wait [0,0];
    if (flight_status = InAir) then
        skill[takeoff].status := success;
        to done
    else
        // Skill 'takeoff' failed success_postcondition 'at_altitude': (flight_status = InAir)
        to failed_success_postcondition_at_altitude
    end

from failed_success_postcondition_at_altitude
    wait [0,0];
    skill[takeoff].status := success;
to done

from failure_grounded
    wait [0,0];
    skill[takeoff].val := takeoff_ret_val(takeoff_failure_grounded);
    // Skill 'takeoff' failure 'grounded'
    skill[takeoff].inv_active := false;
    motion := Free; 
to check_failure_postcondition_grounded

from check_failure_postcondition_grounded
    wait [0,0];
    if (flight_status = OnGround) then
        skill[takeoff].status := failure;
        to done
    else
        // Skill 'takeoff' failed failure_postcondition 'grounded': (flight_status = OnGround)
        to failed_failure_postcondition_grounded
    end

from failed_failure_postcondition_grounded
    wait [0,0];
    skill[takeoff].status := failure;
to done

from failure_emergency
    wait [0,0];
    skill[takeoff].val := takeoff_ret_val(takeoff_failure_emergency);
    // Skill 'takeoff' failure 'emergency'
    skill[takeoff].inv_active := false;
    motion := Free; 
to check_failure_postcondition_emergency

from check_failure_postcondition_emergency
    wait [0,0];
    if (flight_status = InAirUnsafe) then
        skill[takeoff].status := failure;
        to done
    else
        // Skill 'takeoff' failed failure_postcondition 'emergency': (flight_status = InAirUnsafe)
        to failed_failure_postcondition_emergency
    end

from failed_failure_postcondition_emergency
    wait [0,0];
    skill[takeoff].status := failure;
to done

from done
    wait [0,0];
    skill[takeoff].caller := None;
    // Skill 'takeoff' returning control to caller and back to 'ether'
to ether

from error
    wait [0,0];
    // Skill 'takeoff' has an error in its model, check the returned values from real actions
to ether

from ether
    wait [0,0];
to start_
\end{lstlisting}

\section{Complete \hfiacre{} processes for a composite skill}
\label{app:fcsh}

\begin{lstlisting}[caption={The \hfiacre{} main process of the \skill{uav\_mission} composite skill.}, numbers=left, xleftmargin=15pt, label={lst:fcsh}, language=fiacre]
process skill_uav_mission
    (&skill: skill_array, &flight_status: sv_flight_status, &target: sv_target, 
     &mission_status: sv_mission_status, &localization_status: sv_localization_status, 
     &motion: sv_motion, &battery: sv_battery, &camera: sv_camera) is

states start_, check_precondition, precondition_satisfied, precondition_unsatisfied,
    body_branch, NS1, NS1_NS3_sync, NS3, NS5, NS5_NS4_sync, NS4, N3_T, N3_T_NS6_sync,
    NS8, NS6, NS7, NS9, NS10, N2_T, N2_T_NS11_sync, NS12, NS11, N1_T, N1_T_NS13_sync, 
    NS14, NS13, NS15, failure_mission_failed, success_mission_accomplished, done, ether

var ignoreb: bool

from start_
    wait [0,0];
    on (not (skill[uav_mission].caller = None));
    skill[uav_mission].status := no_status;
    // Skill 'uav_mission' has been called
to check_precondition

from check_precondition
    wait [0,0];
    on (not (invariant_active(skill, flight_status, target, mission_status, localization_status, motion, battery, camera)));
    if (true) then
        to precondition_satisfied
    else
        to precondition_unsatisfied
    end

from precondition_unsatisfied
    wait [0,0];
    // Skill 'uav_mission' precondition (true) UNsatisfied
to done

from precondition_satisfied
    wait [0,0];
    // Skill 'uav_mission' precondition (true) is satisfied
    mission_status := Ongoing; 
to body_branch

from body_branch
    wait [0,0];
to NS1 //  

// from: NS1 to: NS3 type: ET_GOAL expr: (start_drone)
from NS1
    wait [0,0];
    // Pass the control to (start_drone)
    // Skill 'uav_mission' calling (start_drone)
    skill[start_drone].caller := uav_mission;
    // synthesized action arg index  0
    skill[start_drone].ArgIndex := 0;
to NS1_NS3_sync

from NS1_NS3_sync
    wait [0,0];
    // Wait the control back from (start_drone)
    on (skill[start_drone].caller = None);
    // Skill 'uav_mission' call to (start_drone) returned
to NS3

// from: NS3 to: NS5 type: ET_GOAL expr: (^ (localization_status Valid))
from NS3
    // Wait on state variable value (localization_status Valid)
    wait [0,0];
    // Skill 'uav_mission' waited on condition (localization_status = Valid)
    on (localization_status = Valid);
to NS5

// from: NS5 to: NS4 type: ET_GOAL expr: (takeoff height 3.0 duration 0)
from NS5
    wait [0,0];
    // Pass the control to (takeoff height 3.0 duration 0)
    // Skill 'uav_mission' calling (takeoff height 3.0 duration 0)
    skill[takeoff].caller := uav_mission;
    // synthesized action arg index  1
    skill[takeoff].ArgIndex := 1;
to NS5_NS4_sync

from NS5_NS4_sync
    wait [0,0];
    // Wait the control back from (takeoff height 3.0 duration 0)
    on (skill[takeoff].caller = None);
    // Skill 'uav_mission' call to (takeoff height 3.0 duration 0) returned
to NS4

// from: NS4 to: N3 type: ET_IF expr: (= takeoff.status success)
from NS4
    wait [0,0];
    // Test on (= takeoff.status success)
    // Skill 'uav_mission' testing expression (= takeoff.status success)
    if (skill[takeoff].status = success) then
        // Skill 'uav_mission' expression (= takeoff.status success) is TRUE
        to N3_T
    else
        // Skill 'uav_mission' expression (= takeoff.status success) is FALSE
        to NS8
    end

// splitting on N3_T, level 1, index 0
from N3_T
    wait [0,0];
    // Pass the parallel control
    skill[uav_mission_branch_1_0].caller := uav_mission;
    skill[uav_mission_branch_1_1].caller := uav_mission;
to N3_T_NS6_sync

from N3_T_NS6_sync
    wait [0,0];
    // wait the control back from all // branches
    on (
        (skill[uav_mission_branch_1_0].caller = None) and
        (skill[uav_mission_branch_1_1].caller = None) and 
        true);
to NS6

// from: NS8 to: NS7 type: ET_GOAL expr: 
from NS8
    wait [0,0];
to NS7

// from: NS6 to: NS9 type: ET_GOAL expr: 
from NS6
    wait [0,0];
to NS9

// from: NS7 to: NS10 type: ET_GOAL expr: (printf "Mission failed")
from NS7
    wait [0,0];
    // will print user trace: "Mission failed"
    // Skill 'uav_mission' prints a user message
to NS10

// from: NS9 to: N2 type: ET_IF expr: (= goto_waypoint.status success)
from NS9
    wait [0,0];
    // Test on (= goto_waypoint.status success)
    // Skill 'uav_mission' testing expression (= goto_waypoint.status success)
    if (skill[goto_waypoint].status = success) then
        // Skill 'uav_mission' expression (= goto_waypoint.status success) is TRUE
        to N2_T
    else
        // Skill 'uav_mission' expression (= goto_waypoint.status success) is FALSE
        to NS12
    end

// from: NS10 to: NS38 type: ET_GOAL expr: (failure mission_failed)
from NS10
    wait [0,0];
    // Skill 'uav_mission' is done executing and returns with failure mission_failed
to failure_mission_failed

// from: N2_T to: NS11 type: ET_GOAL expr: (landing)
from N2_T
    wait [0,0];
    // Pass the control to (landing)
    // Skill 'uav_mission' calling (landing)
    skill[landing].caller := uav_mission;
    // synthesized action arg index  0
    skill[landing].ArgIndex := 0;
to N2_T_NS11_sync

from N2_T_NS11_sync
    wait [0,0];
    // Wait the control back from (landing)
    on (skill[landing].caller = None);
    // Skill 'uav_mission' call to (landing) returned
to NS11

// from: NS12 to: NS8 type: ET_GOAL expr: 
from NS12
    wait [0,0];
to NS8

// from: NS11 to: N1 type: ET_IF expr: (= landing.status success)
from NS11
    wait [0,0];
    // Test on (= landing.status success)
    // Skill 'uav_mission' testing expression (= landing.status success)
    if (skill[landing].status = success) then
        // Skill 'uav_mission' expression (= landing.status success) is TRUE
        to N1_T
    else
        // Skill 'uav_mission' expression (= landing.status success) is FALSE
        to NS14
    end

// from: N1_T to: NS13 type: ET_GOAL expr: (shutdown_drone)
from N1_T
    wait [0,0];
    // Pass the control to (shutdown_drone)
    // Skill 'uav_mission' calling (shutdown_drone)
    skill[shutdown_drone].caller := uav_mission;
    // synthesized action arg index  0
    skill[shutdown_drone].ArgIndex := 0;
to N1_T_NS13_sync

from N1_T_NS13_sync
    wait [0,0];
    // Wait the control back from (shutdown_drone)
    on (skill[shutdown_drone].caller = None);
    // Skill 'uav_mission' call to (shutdown_drone) returned
to NS13

// from: NS14 to: NS12 type: ET_GOAL expr: 
from NS14
    wait [0,0];
to NS12

// from: NS13 to: NS15 type: ET_GOAL expr: (printf "Mission Accomplished")
from NS13
    wait [0,0];
    // will print user trace: "Mission Accomplished"
    // Skill 'uav_mission' prints a user message
to NS15

// from: NS15 to: NS14 type: ET_GOAL expr: (success mission_accomplished)
from NS26
    wait [0,0];
    // Skill 'uav_mission' is done executing and returns with success mission_accomplished
to success_mission_accomplished

from success_mission_accomplished
    wait [0,0];
    skill[uav_mission].val := uav_mission_ret_val(uav_mission_success_mission_accomplished);
    // Skill 'uav_mission' success 'mission_accomplished'
    mission_status := Succeeded; 
    skill[uav_mission].status := success;
to done

from failure_mission_failed
    wait [0,0];
    skill[uav_mission].val := uav_mission_ret_val(uav_mission_failure_mission_failed);
    // Skill 'uav_mission' failure 'mission_failed'
    mission_status := Failed; 
    skill[uav_mission].status := failure;
to done

from done
    wait [0,0];
    skill[uav_mission].caller := None;
    // Skill 'uav_mission' returning control to caller and back to 'ether'
to ether
\end{lstlisting}

\begin{lstlisting}[caption={The \hfiacre{} process (first parallel branch) of the \skill{uav\_mission}.}, numbers=left, xleftmargin=15pt, label={lst:fcshp1}, language=fiacre]
process skill_branch_uav_mission_1_0
      (&skill: skill_array, &flight_status: sv_flight_status, &target: sv_target, 
       &mission_status: sv_mission_status, &localization_status: sv_localization_status, 
       &motion: sv_motion, &battery: sv_battery, &camera: sv_camera) is

states start_, body_branch, N3_T, N3_T_NS6_sync, NS6, done, ether

var ignoreb: bool

from start_
    wait [0,0];
    on (not (skill[uav_mission_branch_1_0].caller = None));
to body_branch

from body_branch
    wait [0,0];
to N3_T //  

// from: N3_T to: NS6 type: ET_GOAL expr: (camera_tracking)
from N3_T
    wait [0,0];
    // Pass the control to (camera_tracking)
    // Skill 'uav_mission' calling (camera_tracking)
    skill[camera_tracking].caller := uav_mission_branch_1_0;
    // synthesized action arg index  0
    skill[camera_tracking].ArgIndex := 0;
to N3_T_NS6_sync

from N3_T_NS6_sync
    wait [0,0];
    // Wait the control back from (camera_tracking)
    on (skill[camera_tracking].caller = None);
    // Skill 'uav_mission' call to (camera_tracking) returned
to NS6

// Joining on NS6, this branch is done
from NS6 //  join 
    wait [0,0];
to done

from done
    wait [0,0];
    skill[uav_mission_branch_1_0].caller := None;
to ether

from ether
    wait [0,0];
to start_
\end{lstlisting}

\begin{lstlisting}[caption={The \hfiacre{} process (second parallel branch) of the \skill{uav\_mission}.}, numbers=left, xleftmargin=15pt, label={lst:fcshp2}, language=fiacre]
process skill_branch_uav_mission_1_1
      (&skill: skill_array, &flight_status: sv_flight_status, &target: sv_target, 
       &mission_status: sv_mission_status, &localization_status: sv_localization_status, 
       &motion: sv_motion, &battery: sv_battery, &camera: sv_camera) is

states start_, body_branch, N3_T, N3_T_NS1_sync, NS1, NS2, NS2_NS5_sync, NS5, NS6, done, ether

var ignoreb: bool

from start_
    wait [0,0];
    on (not (skill[uav_mission_branch_1_1].caller = None));
to body_branch

from body_branch
    wait [0,0];
to N3_T //  

// from: N3_T to: NS1 type: ET_GOAL expr: (goto_waypoint x 1 y 2 z 3 yaw 0 duration 0)
from N3_T
    wait [0,0];
    // Pass the control to (goto_waypoint x 1 y 2 z 3 yaw 0 duration 0)
    // Skill 'uav_mission' calling (goto_waypoint x 1 y 2 z 3 yaw 0 duration 0)
    skill[goto_waypoint].caller := uav_mission_branch_1_1;
    // synthesized action arg index  2
    skill[goto_waypoint].ArgIndex := 2;
to N3_T_NS1_sync

from N3_T_NS1_sync
    wait [0,0];
    // Wait the control back from (goto_waypoint x 1 y 2 z 3 yaw 0 duration 0)
    on (skill[goto_waypoint].caller = None);
    // Skill 'uav_mission' call to (goto_waypoint x 1 y 2 z 3 yaw 0 duration 0) returned
to NS1

// from: NS1 to: NS2 type: ET_GOAL expr: (^ 2)
from NS1
    // Wait 2 seconds
    // Skill 'uav_mission' waited 2 seconds (200 ticks)
    wait [200,200];
to NS13

// from: NS13 to: NS15 type: ET_GOAL expr: (goto_waypoint x -3 y -2 z 4 yaw 1.4 duration 0)
from NS13
    wait [0,0];
    // Pass the control to (goto_waypoint x -3 y -2 z 4 yaw 1.4 duration 0)
    // Skill 'uav_mission' calling (goto_waypoint x -3 y -2 z 4 yaw 1.4 duration 0)
    skill[goto_waypoint].caller := uav_mission_branch_1_1;
    // synthesized action arg index  3
    skill[goto_waypoint].ArgIndex := 3;
to NS13_NS15_sync

from NS13_NS15_sync
    wait [0,0];
    // Wait the control back from (goto_waypoint x -3 y -2 z 4 yaw 1.4 duration 0)
    on (skill[goto_waypoint].caller = None);
    // Skill 'uav_mission' call to (goto_waypoint x -3 y -2 z 4 yaw 1.4 duration 0) returned
to NS15

// from: NS15 to: NS6 type: ET_GOAL expr: (camera_tracking.interrupt)
from NS15
    wait [0,0];
    // will interrupt camera_tracking
    // Skill 'uav_mission' is interrupting camera_tracking
    ignoreb := Fiacre_camera_tracking_interrupt_action();
to NS6

// Joining on NS6, this branch is done
from NS6 //  join 
    wait [0,0];
to done

from done
    wait [0,0];
    skill[uav_mission_branch_1_1].caller := None;
to ether

from ether
    wait [0,0];
to start_
\end{lstlisting}

\begin{lstlisting}[caption={The \hfiacre{} process monitoring under and over shooting of the \skill{uav\_mission} skill.}, numbers=left, xleftmargin=15pt, label={lst:fcshp3}, language=fiacre]
process skill_uav_mission_watchdog(&skill: skill_array) is 

states start_, monitor, skill_overshoot, skill_not_undershoot

var ignorep: nat, overshoot, undershoot : bool

from start_
    wait [0,0];
    on (not (skill[uav_mission].caller = None)); // Skill 'uav_mission' watchdog started
    overshoot := false;
    undershoot := true;
to monitor

from monitor
    select
        on (skill[uav_mission].caller = None);
        if undershoot then // execution terminated before the undershoot elapsed
            // WARNING: Skill 'uav_mission' undershoot 60 s (6000 ticks)
            null
        end;
        to start_
    [] // undershoot
        on undershoot;
        wait [6000, 6000]; // wait the undershoot time value
        to skill_not_undershoot
    [] // overshoot
        on not undershoot and not overshoot;
        wait [6000, 6000]; // we already waited 600 for undershoot + 600 = 1200
        // WARNING: Skill 'uav_mission' overshoot 120 s (12000 ticks) 
        to skill_overshoot
    end

from skill_overshoot
    wait [0,0];
    overshoot := true;
to monitor

from skill_not_undershoot // we got here before the execution terminated
    wait [0,0];
    undershoot := false; // hence we did NOT undershoot
to monitor
\end{lstlisting}

\end{document}